\documentclass[journal]{IEEEtran}

\usepackage{graphicx}
\usepackage{booktabs}
\usepackage{subcaption}
\usepackage{amssymb}
\usepackage{multirow}
\usepackage{dsfont }
\usepackage{xspace}
\usepackage{amsmath}
\usepackage{cite}
\usepackage{hyperref}

\usepackage{adjustbox}
\usepackage{color}
\usepackage{mathtools}
\usepackage{authblk}

\ifCLASSINFOpdf
\else
\fi

\begin{document}
\title{Uncertainty Awareness on Unsupervised Domain Adaptation for Time Series Data}

\author{Weide~Liu,
        Xiaoyang Zhong,
        Lu Wang,
        Jingwen Hou,
        Yuemei Luo,
        Jiebin Yan,
        and Yuming Fang

\thanks{Weide Liu is with College of Computing and Data Science, Nanyang Technological University (NTU), Singapore 639798 (e-mail: weide001@e.ntu.edu.sg).}

\thanks{X Zhong, J Hou, J Yan and Y Fang are with School of Computing and Artificial Intelligence, Jiangxi University of Finance and Economics, China,
email: zhongaobo429@foxmail.com, jingwen003@e.ntu.edu.sg, jiebinyan@foxmail.com, fa0001ng@e.ntu.edu.sg }

\thanks{L Wang is with Institute for Infocomm Research (I\textsuperscript{2}R), A*STAR, Singapore,
email: wang\_lu@i2r.a-star.edu.sg}

\thanks{Y Luo is with School of Artificial Intelligence, Nanjing University of Information Science and Technology, China,
email: luoy0021@e.ntu.edu.sg}

\thanks{Corresponding author: Lu Wang. (e-mail: wang\_lu@i2r.a-star.edu.sg)}
}

% The paper headers
\markboth{}%
{Shell \MakeLowercase{\textit{et al.}}: Bare Demo of IEEEtran.cls for IEEE Journals}

% make the title area
\maketitle

\begin{abstract}
Unsupervised domain adaptation methods seek to generalize effectively on unlabeled test data, especially when encountering the common challenge in time series data that distribution shifts occur between training and testing datasets.
In this paper, we propose incorporating multi-scale feature extraction and uncertainty estimation to improve the model’s generalization and robustness across domains. Our approach begins with a multi-scale mixed input architecture that captures features at different scales, increasing training diversity and reducing feature discrepancies between the training and testing domains. 
Based on the mixed input architecture, we further introduce an uncertainty awareness mechanism based on evidential learning by imposing a Dirichlet prior on the labels to facilitate both target prediction and uncertainty estimation.
The uncertainty awareness mechanism enhances domain adaptation by aligning features with the same labels across different domains, which leads to significant performance improvements in the target domain. 
Additionally, our uncertainty-aware model demonstrates a much lower Expected Calibration Error (ECE), indicating better-calibrated prediction confidence.
Our experimental results show that this combined approach of mixed input architecture with the uncertainty awareness mechanism achieves state-of-the-art performance across multiple benchmark datasets, underscoring its effectiveness in unsupervised domain adaptation for time series data.
Our code is available at \href{https://github.com/ZhongAobo/Evidential-HAR}{https://github.com/ZhongAobo/Evidential-HAR}.
\end{abstract}


\IEEEpeerreviewmaketitle

\section{Introduction}

 \begin{figure*}[t]
  \centering
    \includegraphics[width=0.9\linewidth]{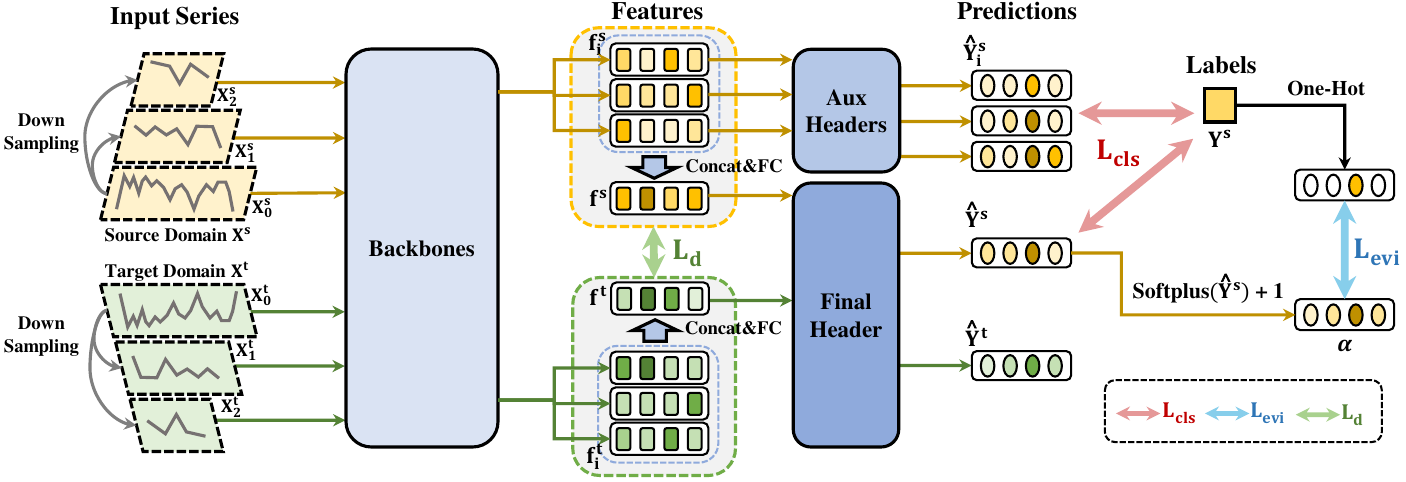}
    \caption{The architecture of our method. Where $L_{d}$ is domain loss, $L_{cls}$ is classifier loss, and $L_{evi}$ is our evidential loss. $f^s$ and $f^s_i$ are features of the images for the source domain and source domain at scale $i$. “Aux Headers” refers to the auxiliary classification heads for different scales and “Final Header” is the final classification head using all concatenated features of different scales.}
    \label{Figure:pipeline}
\end{figure*}
In various real-world applications such as healthcare and manufacturing, classifying time series data is increasingly critical. 
Deep learning, in particular, has been gaining attention for its ability to learn the temporal dynamics embedded in complex data patterns, assuming an abundance of labeled data is available~\cite{ismail2019deep}. 
However, labeling this data is often complicated and demanding because it requires specialized domain knowledge~\cite{chang2020systematic}. 
One approach to alleviate the burden of labeling involves using annotated data from a related domain that is already labeled or easier to label, known as the source domain, to train the model, which is then tested on a different target domain of interest. 
This strategy can be problematic due to the potentially distinct data distributions between the source and target domains, which might result in significant domain shifts that compromise the model's performance on the target domain. This challenge is common in time series classification tasks such as Human Activity Recognition (HAR)~\cite{har_data} and Sleep Stage Classification (SSC)~\cite{eldele2022adast}. Specifically, Human Activity Recognition (HAR) entails detecting human activities through data collected from embedded sensors in devices like smartphones and wearables, or sensors installed in the environment. HAR has wide-ranging applications, including health monitoring, elderly care, sports training, and interactive games~\cite{har_data}.
Similarly, sleep stage classification (SSC) classifies various sleep stages using physiological signals, which is essential for diagnosing sleep disorders and analyzing sleep patterns~\cite{sleepEDF_dataset}. However, variations in sensor placement or patient physiology can introduce distribution mismatches that hinder generalization~\cite{eldele2022adast}.

Differences in temporal characteristics and operational conditions between training and testing datasets often lead to a marked decline in the performance of deep learning models. This challenge, known as the domain shift problem, undermines the efficacy of deep learning in real-world time series applications.
Unsupervised Domain Adaptation (UDA) aims to mitigate this issue by adapting models trained on a labeled source domain to perform effectively on an unlabeled, shifted target domain. While there is extensive research on UDA for visual tasks \cite{tl_survey,visual_da,wu2022multiple}, its application to time series data remains comparatively underexplored.

Current methodologies for time series domain adaptation typically align source and target domains by minimizing statistical discrepancies (e.g., via distribution matching \cite{deepcoral,HoMM,ddc,dsan}) or leveraging adversarial training with domain discriminators \cite{Ganin2016,CDAN,slarda,ijcai2021p378}. While these approaches encourage feature robustness across domains, they often overlook a critical factor: model uncertainty. In real-world scenarios, domain shifts introduce ambiguity in how well-aligned features generalize to the target domain. Without explicitly quantifying uncertainty, models risk overconfident predictions on poorly adapted regions of the target data, leading to unreliable performance. Uncertainty estimation thus becomes indispensable for evaluating a model’s generalization capacity and identifying adaptation gaps caused by temporal variations in sensor conditions.

To address these challenges, we propose a novel unsupervised domain adaptation (UDA) framework that integrates multi-scale temporal modeling with evidential uncertainty-aware learning. First, our mixed multi-scale architecture captures hierarchical temporal patterns (e.g., short-term fluctuations and long-term trends), enabling robust feature extraction for domain-varying signals. By processing data across scales, the model reduces ambiguity in temporal representations, critical for adaptation tasks where domain shifts occur at specific frequencies. Second, we incorporate evidential learning \cite{2018Evidential,deep_evidential} to estimate epistemic uncertainty, which quantifies prediction confidence. Unlike traditional probabilistic methods, evidential learning models uncertainty through evidence accumulation, treating network outputs as parameters of a Dirichlet distribution (conjugate to the categorical distribution). This approach avoids reliance on softmax-based point estimates and optimizes predictions via an evidential Bayesian risk loss, directly reflecting uncertainty during domain transfer.

The synergy of these components ensures robust alignment between domains while identifying samples of high uncertainty (e.g., irregular patterns or unseen dynamics) where adaptation may falter. By prioritizing uncertain samples during training, the framework refines feature representations, enhancing robustness against temporal shifts. For instance, in applications like HAR or SSC, where sensor heterogeneity or physiological variability induce uncertainty, the model adapts cautiously to ambiguous target data, avoiding overconfidence. Experiments across five time series datasets demonstrate that our method achieves state-of-the-art performance, balancing predictive accuracy with interpretable uncertainty quantification for reliable cross-domain adaptation.
 \begin{figure*}[t]
  \centering
    \includegraphics[width=0.9\linewidth]{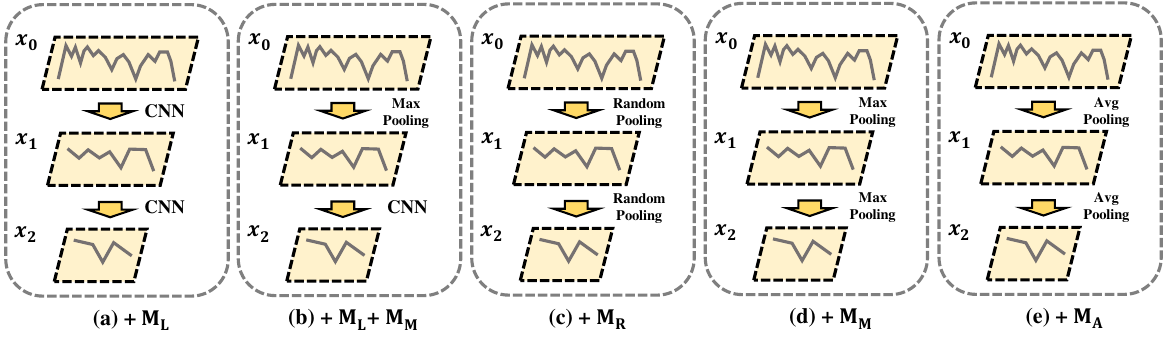}
    \caption{Different down-sampling method in our multi-scale mixing architecture, where (a) denotes down-sampling with the 1D CNN, (b)  denotes down-sampling with the 1D CNN and max-pooling, (c) denotes down-sampling with the random-pooling, (d) denotes down-sampling with the max-pooling and (e) denotes down-sampling with the avg-pooling.
    }
    \label{Figure:pooling}
\end{figure*}
\begin{table*}[ht]
\caption{The table presents a comparison of our method with other approaches, where the performance of the baseline model is benchmarked against AdaTime \protect\cite{adatime}. Uncertainty is imposed by two different priors, i.e., the normal-inverse-gamma (NIG) distribution with normal likelihood \protect\cite{deep_evidential} and the Dirichlet (DIR) distribution with categorical likelihood \protect\cite{2018Evidential}. The $ml$, $ce$, $mse$ correspond to 3 different losses according to Equation~(\ref{eq_typeII}), Equation~(\ref{eq_ce}) and Equation~(\ref{eq_mse}), respectively. The Bayesian risk loss with cross-entropy shows the best f1 score on average.}

\renewcommand\arraystretch{1} 
\begin{adjustbox}{width=0.9\textwidth,center}
\begin{tabular}{c|l|ccccccccccccc|c}
\toprule
Dataset&Method&noAdapt&DDC&Deep&HoMM&DANN&MMDA&DSAN&CDAN&DIRT&CoDATS&AdvSKM&SASA&{CLUDA}&AVG \\ \hline
\multirow{5}{*}{UCIHAR} 
&Baseline&65.94 &82.29 &86.30 &88.52 &88.26 &89.39 &91.46 &90.72 &93.68 &88.20 &80.10 &85.00 &91.47&86.26  \\
&+NIG &70.34 &87.61 &89.33 &87.63 &87.72 &88.56 &92.23 &88.57 &91.36 &88.11 &87.01 &83.87 &90.98&87.26  \\
&+DIR$_{ml}$&\textbf{82.69}&\textbf{88.61}&89.86&91.77&89.58&\textbf{90.31}&90.51&90.40&92.90&91.33&88.14&87.55&91.88&89.66  \\
&+DIR$_{mse}$&78.40&88.15&\textbf{90.07}&\textbf{92.11}&\textbf{89.76}&90.28&92.36&90.51&93.12&\textbf{92.87}&\textbf{88.39}&86.40&91.57&89.54  \\
&+DIR$_{ce}$&80.99&88.49&89.61&91.77&89.55&89.98&\textbf{92.43}&\textbf{91.79}&\textbf{93.72}&91.97&88.05&\textbf{88.37}&\textbf{91.97}&\textbf{89.90}  \\ \hline
\multirow{5}{*}{WISDM} 
&Baseline&48.60 &53.78 &54.19 &56.92 &62.41 &59.82 &61.08 &55.59 &59.59 &62.06 &54.82 &53.34 &57.36&56.89  \\
&+NIG &50.01 &53.57 &53.50 &51.01 &57.02 &60.68 &57.62 &52.96 &59.46 &58.67 &53.61 &54.05 &57.31&55.34  \\
&+DIR$_{ml}$&\textbf{55.91}&58.03&58.21&57.72&\textbf{66.00}&61.15&63.12&56.92&60.20&62.11&51.95&53.17&\textbf{57.75}&58.63  \\
&+DIR$_{mse}$&55.56&\textbf{58.34}&\textbf{58.46}&60.12&60.01&\textbf{61.84}&\textbf{64.23}&\textbf{61.05}&60.65&62.25&55.09&53.50&57.65&59.13  \\
&+DIR$_{ce}$&53.46&57.52&\textbf{58.46}&\textbf{60.18}&62.76&61.22&61.71&59.56&\textbf{61.03}&\textbf{63.93}&\textbf{58.33}&\textbf{54.57}&57.41&\textbf{59.24}  \\ \hline
\multirow{5}{*}{HHAR} 
&Baseline&63.07 &68.29 &71.75 &73.69 &77.89 &76.93 &79.27 &79.07 &80.47 &76.41 &65.88 &75.76 &78.12&74.35  \\
&+NIG &64.55 &69.99 &70.25 &71.54 &78.86 &76.10 &77.93 &72.82 &80.81 &74.84 &67.28 &77.37 &77.12&73.80  \\
&+DIR$_{ml}$&\textbf{68.08}&\textbf{70.51}&\textbf{73.27}&76.15&\textbf{80.15}&79.67&78.92&78.81&81.46&\textbf{79.58}&68.81&76.50&78.13&76.16  \\
&+DIR$_{mse}$&66.87&69.12&73.18&75.99&79.68&\textbf{79.73}&79.29&78.12&\textbf{82.91}&78.62&68.71&76.16&78.31&75.90  \\
&+DIR$_{ce}$&67.84&70.20&73.26&\textbf{76.23}&79.78&78.08&\textbf{79.98}&\textbf{79.87}&82.39&78.84&\textbf{69.10}&\textbf{77.67}&\textbf{78.65}&\textbf{76.30} 
 \\ \hline
\multirow{5}{*}{SSC} 
&Baseline           &51.67&60.88&61.05&\textbf{60.81}&\textbf{60.80}&63.47&59.51&59.51&\textbf{61.38}&57.32&61.18&\textbf{59.81}&59.59&59.77   \\
&+NIG               &51.52&\textbf{61.24}&60.39&60.48&60.50&\textbf{63.99}&59.69&59.32&61.22&56.54&61.14&59.19&59.59&59.60   \\
&+DIR$_{ml}$        &54.87&59.73&61.24&59.29&58.58&63.58&59.06&62.14&58.73&61.69&60.81&58.70&59.65&59.85   \\
&+DIR$_{mse}$       &\textbf{55.30}&60.29&\textbf{61.54}&59.57&58.81&63.03&\textbf{60.24}&\textbf{62.27}&58.80&61.84&\textbf{61.46}&58.71&59.75&60.12   \\
&+DIR$_{ce}$  &54.88&60.02&61.29&60.05&59.39&63.43&59.78&61.98&58.59&\textbf{62.15}&61.18&59.40&\textbf{59.85}&\textbf{60.15}   \\ \hline
\multirow{5}{*}{MFD} 
&Baseline           &72.51&81.54&80.80&81.18&84.06&85.44&81.65&84.64&92.81&84.20&81.47&78.94&92.22&83.19   \\
&+NIG               &73.44&83.59&80.82&83.28&85.93&85.75&82.54&86.60&\textbf{94.46}&83.99&83.70&78.52&91.66&84.17   \\
&+DIR$_{ml}$        &80.90&89.64&88.80&87.06&\textbf{96.52}&97.75&90.66&87.77&89.57&90.42&88.86&90.61&\textbf{93.12}&90.13  \\
&+DIR$_{mse}$       &81.00&89.80&89.06&87.51&96.21&\textbf{97.87}&90.57&87.04&89.98&90.11&\textbf{89.04}&90.74&92.34&90.10   \\
&+DIR$_{ce}$  &\textbf{81.34}&\textbf{90.53}&\textbf{89.53}&\textbf{87.71}&96.32&97.56&\textbf{90.95}&\textbf{87.80}&90.04&\textbf{90.88}&88.74&\textbf{91.12}&93.08&\textbf{90.43}   \\ \hline

\bottomrule
\end{tabular}
\end{adjustbox}
\label{table:sota}
\end{table*}  
 \begin{figure*}[ht]
  \centering
    \includegraphics[width=1
    \linewidth]{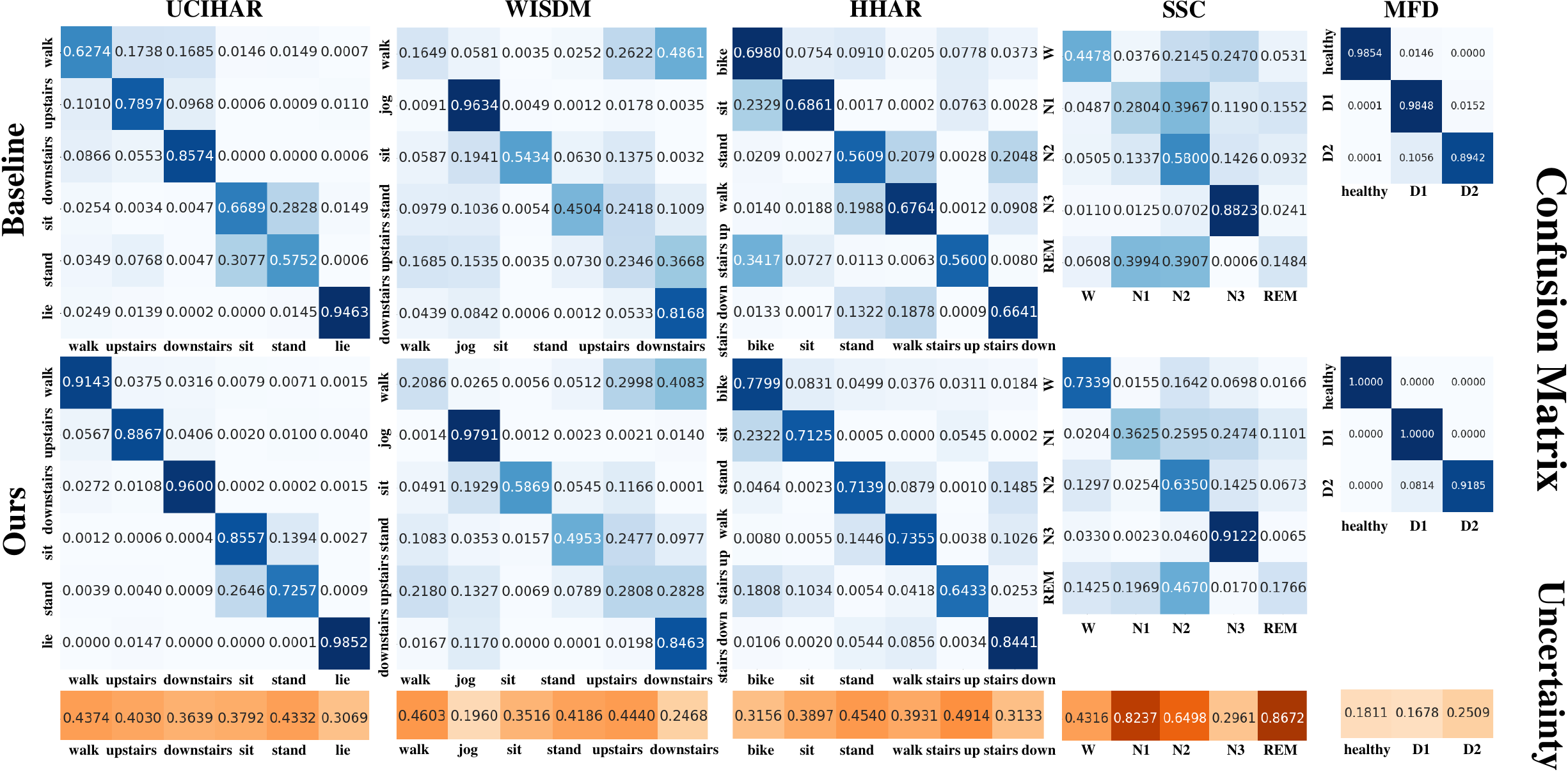}
    \caption{The confusion matrix and uncertainty. Using the noAdapt method as the baseline, test on 5 different datasets.}
    \label{Figure:matrix_uncertainty}
\end{figure*}
 \begin{figure*}[ht]
  \centering
    \includegraphics[width=\textwidth]{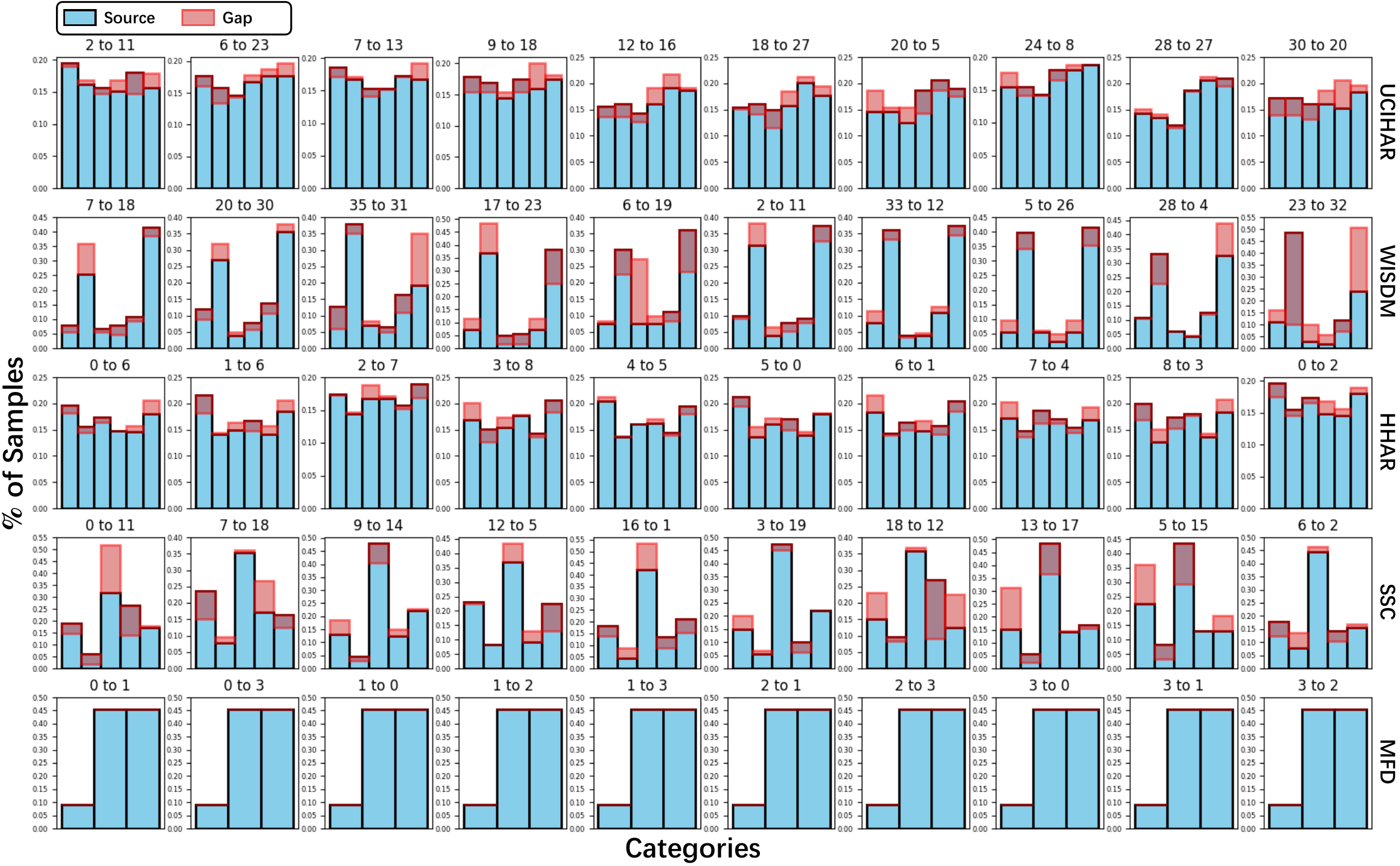}
    \caption{This figure illustrates the label distribution of source data across various dataset splits, along with the gap (indicated by red) between the distributions of source and target data. The $x$ axis represents the different categories, while the $y$ axis represents the distribution density. {
As shown in the figure, we demonstrate the variations in sample quantities across different categories within five datasets and also emphasize the discrepancies in category distributions between the source and target domains.
}}
    \label{Figure:dist}
\end{figure*}
 \begin{figure*}[ht]
  \centering
    \includegraphics[width=1\linewidth]{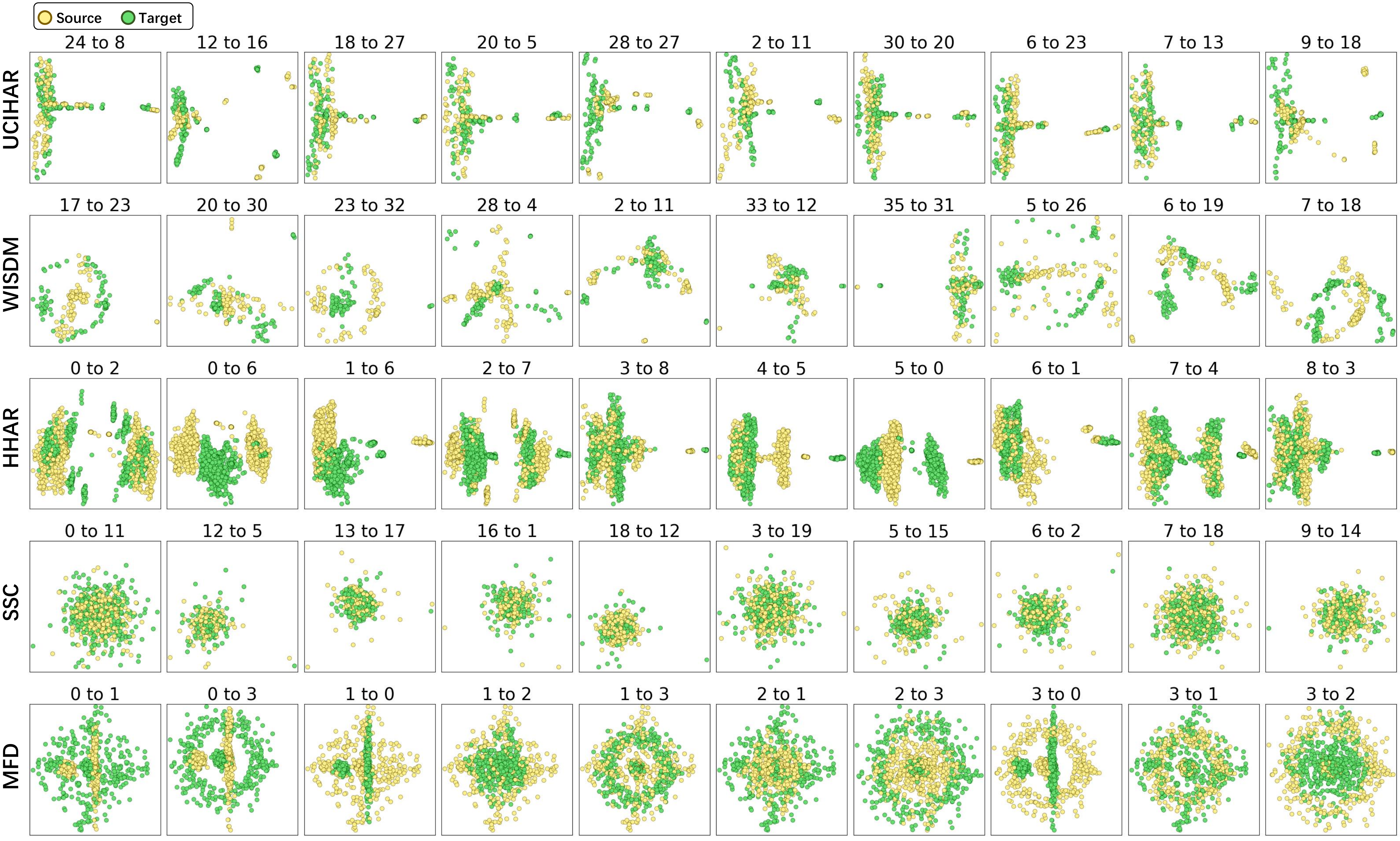}
    \caption{The PCA visualization of samples of different domains across 10 splits (source  and target samples are shown in yellow and green). Distribution shifts are evident between the source and target samples.}
    \label{Figure:all_pca}
\end{figure*}
\begin{table*}[ht]
\caption{The table presents a comparison of our method with other approaches. Baseline* denotes the results reproduced according to the published code provided by the author.
The + UN denotes with the uncertainty awareness, while + $M_{L}$ denotes the multi-scale mixing architecture with learnable parameter (1D CNN), + $M_{R}$ denotes the multi-scale mixing architecture with random pooling, + $M_{M}$ denotes the multi-scale mixing architecture with max pooling, + $M_{A}$ denotes the multi-scale mixing architecture with averaging pooling (detailed architectures are illustrated in Figure~\ref{Figure:pooling}).
}

\renewcommand\arraystretch{1} 
\begin{adjustbox}{width=1\textwidth,center}
\begin{tabular}{c|l|ccccccccccccc|c}
\toprule
Dataset&Method&noAdapt&DDC&Deep&HoMM&DANN&MMDA&DSAN&CDAN&DIRT&CoDATS&AdvSKM&SASA&\textcolor{red}{CLUDA}&AVG \\ \hline
\multirow{8}{*}{UCIHAR} 
&Baseline       &65.94&82.29&86.30&88.52&88.26&89.39&91.46&90.72&93.68&88.20&80.10&85.00&-&85.82    \\
&Baseline*      &70.77&87.55&88.75&90.77&88.36&89.04&91.11&88.20&91.27&88.95&86.74&87.89&91.47&87.76 \\
&+ UN     &80.90&88.49&89.61&91.77&89.55&89.98&92.43&91.79&\textbf{93.72}&91.97&88.05&88.37&91.97&89.90    \\
&+ UN + $M_{L}$      &81.32&90.13&\textbf{90.82}&92.38&91.16&91.07&92.53&92.72&91.32&92.19&90.51 &88.77 &91.47 &90.49    \\
&+ UN + $M_{L}$ + $M_{M}$ &80.02&91.63&90.19&\textbf{92.96}&92.21&91.85&93.22&93.16&91.45&92.03&90.89&\textbf{89.11}&91.44&90.86    \\
&+ UN + $M_{R}$   &80.13&92.14&90.21&92.31&92.05&92.45&92.79&\textbf{93.82}&92.00&91.58&92.28&86.46&90.88&90.70    \\
&+ UN + $M_{M}$      &81.87&\textbf{92.40}&90.76&92.55&\textbf{92.75}&\textbf{92.56}&\textbf{93.55}&93.67&92.64&93.37&\textbf{92.51}&86.87&\textbf{92.03}&\textbf{91.35}   \\
&+ UN + $M_{A}$      &81.04&92.19&90.72&92.40&92.25&92.51&93.51&93.61&91.85&\textbf{93.88}&92.06&88.56&91.99&91.27    \\  \hline

\multirow{8}{*}{WISDM} 
&Baseline       &48.60&53.78&54.19&56.92&62.41&59.82&61.08&55.59&59.59&62.06&54.82&53.34&-&56.85    \\
&Baseline*      &47.90&56.41&55.90&53.27&60.63&55.63&59.64&57.79&57.71&62.21&57.92&53.76&57.36&56.63   \\
&+ UN     &53.46&57.52&58.46&60.18&62.76&61.22&61.71&59.56&61.03&63.93&58.33&54.57&57.41&59.24   \\
&+ UN + $M_{L}$      &\textbf{54.49}&62.19&62.12&\textbf{65.66}&67.88&63.29&66.29&\textbf{62.50}&59.26&\textbf{64.48}&62.74&57.97&57.99&\textbf{62.07}   \\
&+ UN + $M_{L}$ + $M_{M}$ &53.66&61.13&62.99&62.56&67.26&63.58&66.32&59.96&\textbf{61.37}&63.95&64.87&55.64&59.88&61.78   \\
&+ UN + $M_{R}$   &53.57&61.79&\textbf{64.97}&61.34&66.10&63.98&\textbf{66.82}&59.50&59.90&63.65&\textbf{65.13}&55.22&60.83&61.76   \\
&+ UN + $M_{M}$      &52.98&60.92&63.24&60.46&66.64&63.82&64.59&58.51&59.12&60.93&61.58&\textbf{58.98}&\textbf{64.65}&61.26   \\
&+ UN + $M_{A}$      &54.02&\textbf{63.35}&64.68&60.23&\textbf{69.98}&\textbf{64.95}&63.43&58.16&61.33&61.14&61.47&56.07&60.02&61.45   \\  \hline

\multirow{8}{*}{HHAR} 
&Baseline       &63.07&68.29&71.75&73.69&77.89&76.93&79.27&79.07&80.47&76.41&65.88&75.76&-&74.04   \\
&Baseline*      &62.01&66.97&72.98&75.95&78.56&76.51&77.13&78.10&81.27&74.27&65.91&75.90&78.12&74.13   \\
&+ UN     &67.84&70.20&73.26&76.23&79.78&78.08&\textbf{79.98}&79.87&\textbf{82.39}&\textbf{78.84}&69.10&\textbf{77.67}&78.65&76.30   \\
&+ UN + $M_{L}$      &68.99&70.67&76.51&77.84&79.36&81.64&77.30&79.77&80.51&78.01&69.91&77.02&78.88&76.65  \\
&+ UN + $M_{L}$ + $M_{M}$ &69.21&70.74&\textbf{76.54}&78.97&78.63&\textbf{81.87}&79.25&79.48&81.02&76.47&69.98&77.03&78.15&76.72   \\
&+ UN + $M_{R}$   &68.78&70.39&76.14&78.65&78.75&80.85&76.99&80.15&81.21&76.74&70.26&76.34&78.92&76.47   \\
&+ UN + $M_{M}$      &\textbf{69.32}&69.66&75.87&\textbf{79.71}&\textbf{81.63}&80.54&77.45&\textbf{80.77}&81.33&78.24&\textbf{70.59}&76.48&\textbf{79.32}&\textbf{76.99}   \\
&+ UN + $M_{A}$      &67.99&\textbf{71.26}&76.15&78.30&79.27&80.76&78.23&79.92&81.42&76.48&69.35&77.07&77.96&76.47  \\  \hline

\multirow{8}{*}{SSC} 
&Baseline       &51.67&60.88&61.05&60.81&60.80&\textbf{63.47}&59.51&59.51&\textbf{61.38}&57.32&61.18&59.81&-&59.78   \\
&Baseline*      &51.41&59.84&61.00&59.27&59.23&63.13&59.52&60.36&57.54&61.93&60.07&58.86&59.59&59.37   \\
&+ UN     &54.88&60.02&61.29&60.05&59.39&63.43&59.78&61.98&58.59&62.15&61.18&59.40&59.85&60.15   \\
&+ UN + $M_{L}$      &56.21&\textbf{62.86}&63.25&\textbf{63.11}&63.25&62.54&63.00&62.29&59.85&\textbf{63.83}&\textbf{61.98}&60.28&\textbf{60.12}&\textbf{61.74}  \\
&+ UN + $M_{L}$ + $M_{M}$ &56.10&61.88&62.97&61.67&61.40&63.07&62.36&62.39&60.56&61.75&59.43&59.59&59.87&61.00  \\
&+ UN + $M_{R}$   &55.61&61.04&\textbf{63.53}&62.81&\textbf{62.17}&62.62&62.28&\textbf{63.36}&59.69&62.72&61.34&60.88&58.99&61.31   \\
&+ UN + $M_{M}$      &\textbf{56.24}&61.84&63.07&61.98&61.56&62.06&\textbf{63.46}&62.56&60.13&62.69&60.97&\textbf{61.03}&59.47&61.31   \\
&+ UN + $M_{A}$      &54.78&59.82&63.51&61.04&61.28&62.31&62.84&61.14&60.11&62.18&60.46&60.38&59.45&60.72   \\  \hline

\multirow{8}{*}{MFD} 
&Baseline       &72.51&81.54&80.80&81.18&84.06&85.44&81.65&84.64&92.81&84.20&81.47&78.94&-&82.44   \\
&Baseline*      &77.23&88.08&88.19&86.72&96.30&97.27&85.94&87.67&88.51&87.35&88.72&90.17&92.22&88.80   \\
&+ UN     &81.34&90.53&89.53&87.71&96.32&\textbf{97.56}&90.95&87.80&90.04&90.88&88.74&91.12&93.08&90.43   \\
    &+ UN + $M_{L}$      &80.99&89.36&94.49&91.72&95.62&94.09&90.64&\textbf{95.56}&95.99&88.40&89.77&89.36&\textbf{94.02}&91.54   \\
&+ UN + $M_{L}$ + $M_{M}$ &82.03&\textbf{91.33}&93.47&\textbf{92.67}&95.27&95.09&\textbf{92.92}&91.44&\textbf{96.08}&\textbf{92.06}&90.32&\textbf{92.91}&93.02&\textbf{92.20}   \\
&+ UN + $M_{R}$   &80.45&88.70&94.88&91.78&96.52&95.43&90.89&95.33&90.05&89.16&\textbf{91.58}&88.17&91.98&91.15  \\
&+ UN + $M_{M}$      &81.23&90.61&93.71&91.03&96.24&94.73&91.47&93.61&94.61&90.08&90.58&89.99&93.33&91.63   \\
&+ UN + $M_{A}$      &\textbf{81.88}&91.16&\textbf{95.96}&92.14&\textbf{97.20}&96.47&92.68&91.61&91.69&88.78&89.08&90.21&92.58&91.65   \\  \hline

\bottomrule
\end{tabular}
\end{adjustbox}
\label{table:new_sota}
\end{table*}  
 \begin{figure*}[ht]
  \centering
    \includegraphics[width=0.95\linewidth]{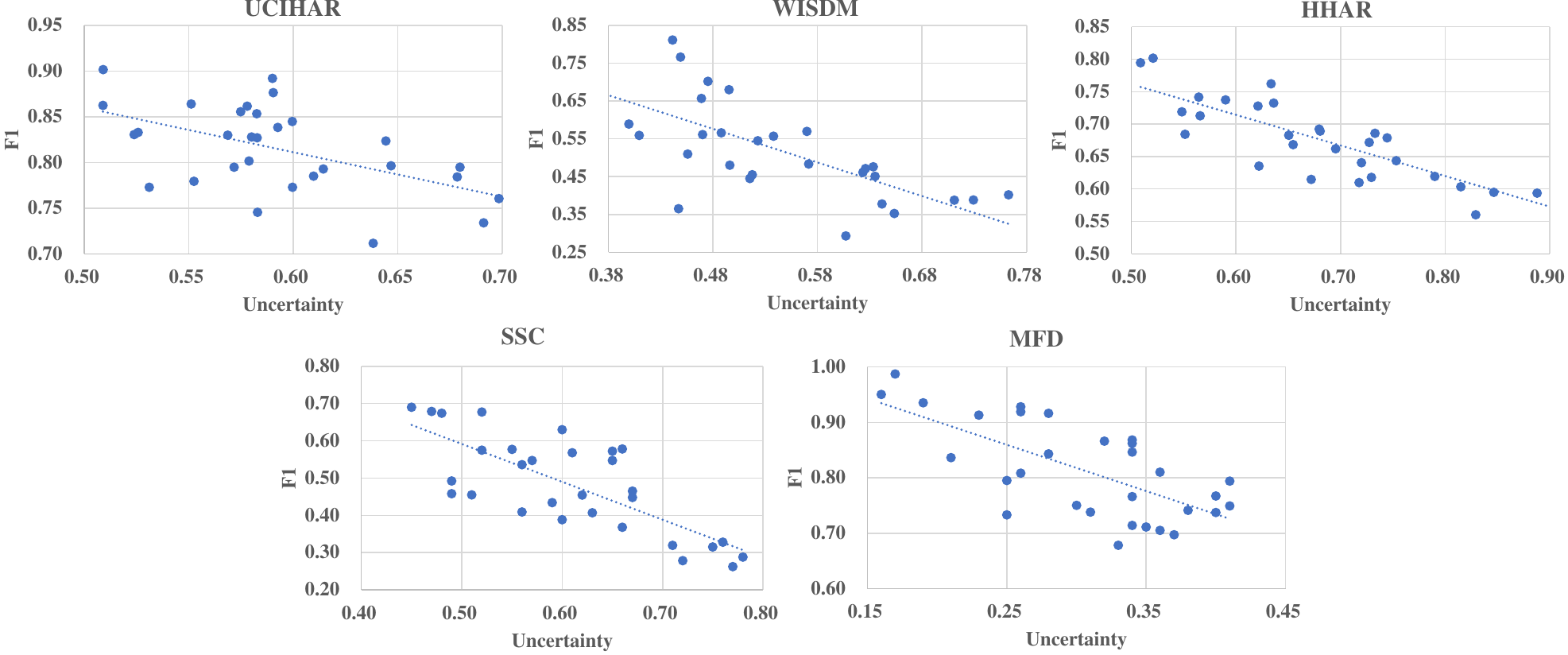}
    \caption{The correlation between F1 score and uncertainty. The baseline is the basic model without domain adaption, randomly select 10 folds on UCIHAR, WISDM, HHAR, SSC and MFD datasets for test, repeat 30 times.}
    \label{Figure:f1_uncertainty}
\end{figure*}
 \begin{figure*}[ht]
  \centering
    \includegraphics[width=1\textwidth]{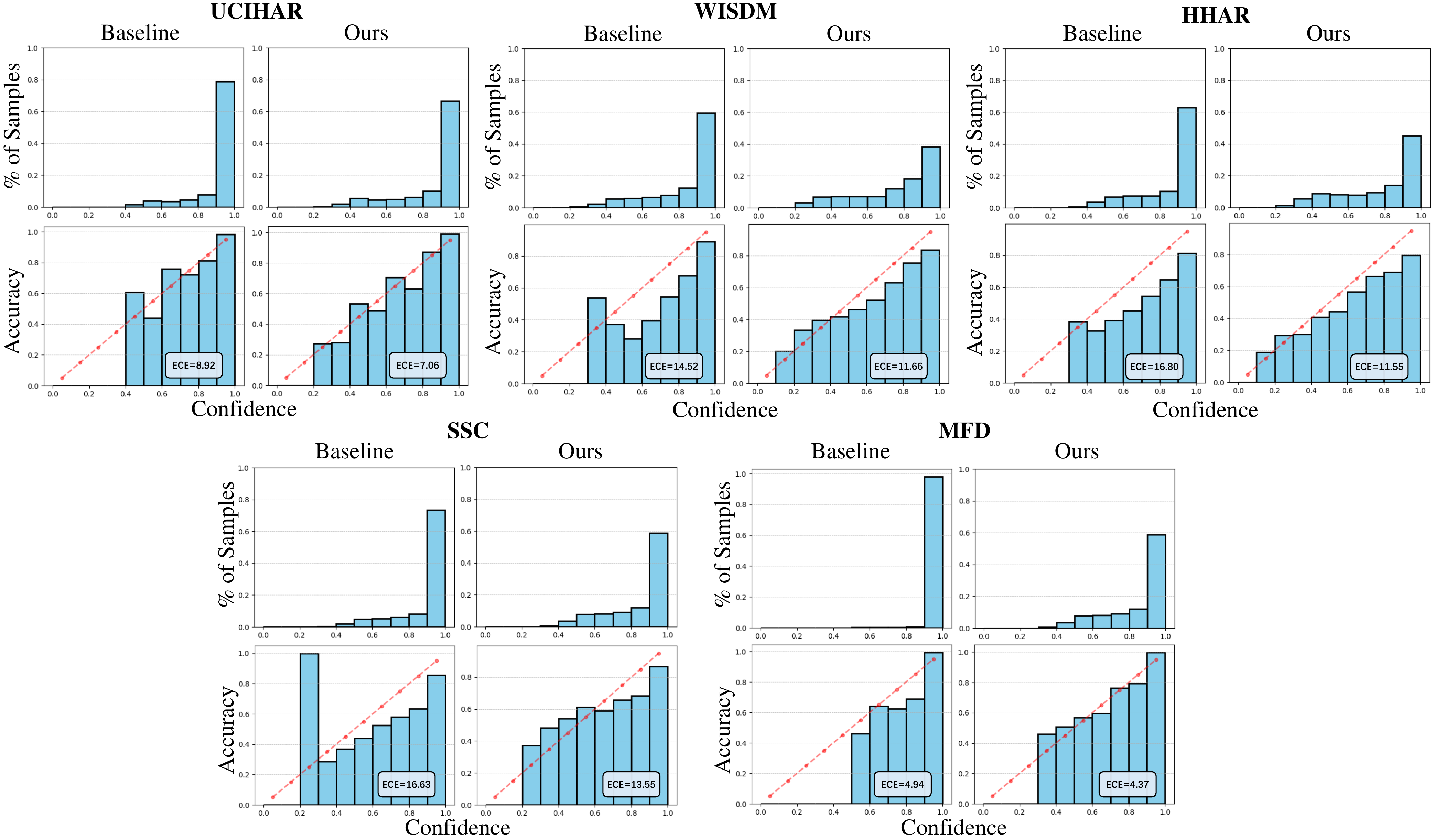}
    \caption{The Expected Calibration Error (ECE) metric. The DDC method served as a baseline for experiments on five datasets. Each subplot's first row indicates the proportion of samples at different confidence levels predicted by the model relative to the total number of samples. The second row illustrates the discrepancy between the classification accuracy (blue bars) and the calibration probability (red dots) for samples at various confidence levels.}
    \label{Figure:ece}
\end{figure*}
 \begin{figure*}[ht]
  \centering
    \includegraphics[width=1\linewidth]{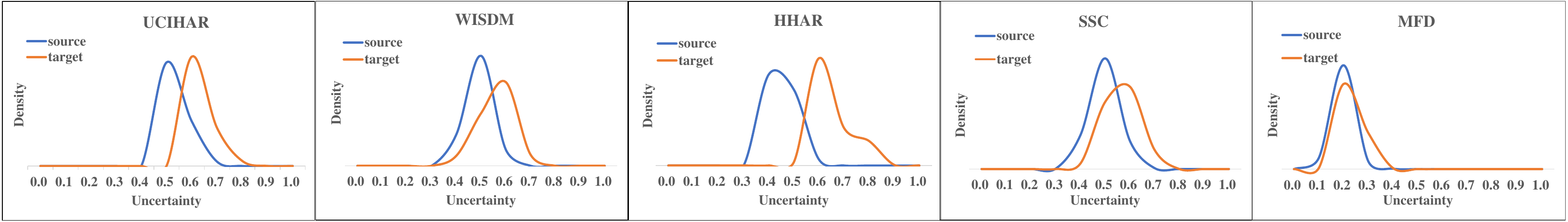}
    \caption{Density of uncertainty. The baseline is the basic model without domain adaption, and add evidential to it to estimate uncertainty on both the source domain and target domain of UCIHAR, WISDM, HHAR, SSC and MFD datasets (data configuration is the same as AdaTime).}
    \label{Figure:uncertainty_density}
\end{figure*}
 \begin{figure*}[t]
  \centering
    \includegraphics[width=1\linewidth]{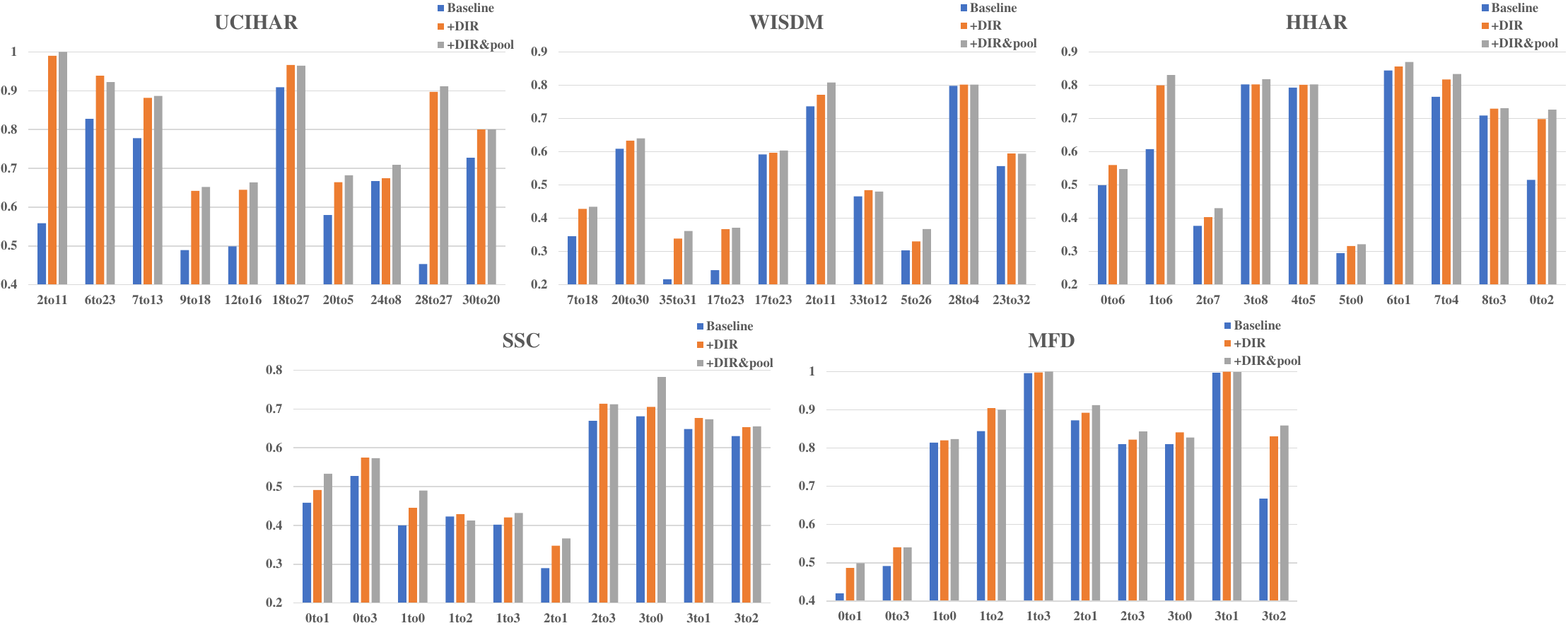}
    \caption{Performance evaluation across different folds over 5 different datasets. The baseline is the basic model without domain adaption. Consistent performance improvements over different folds are found for all datasets.}
    \label{Figure:detail}
\end{figure*}
 \begin{figure*}[ht]
  \centering
    \includegraphics[width=0.6\linewidth]{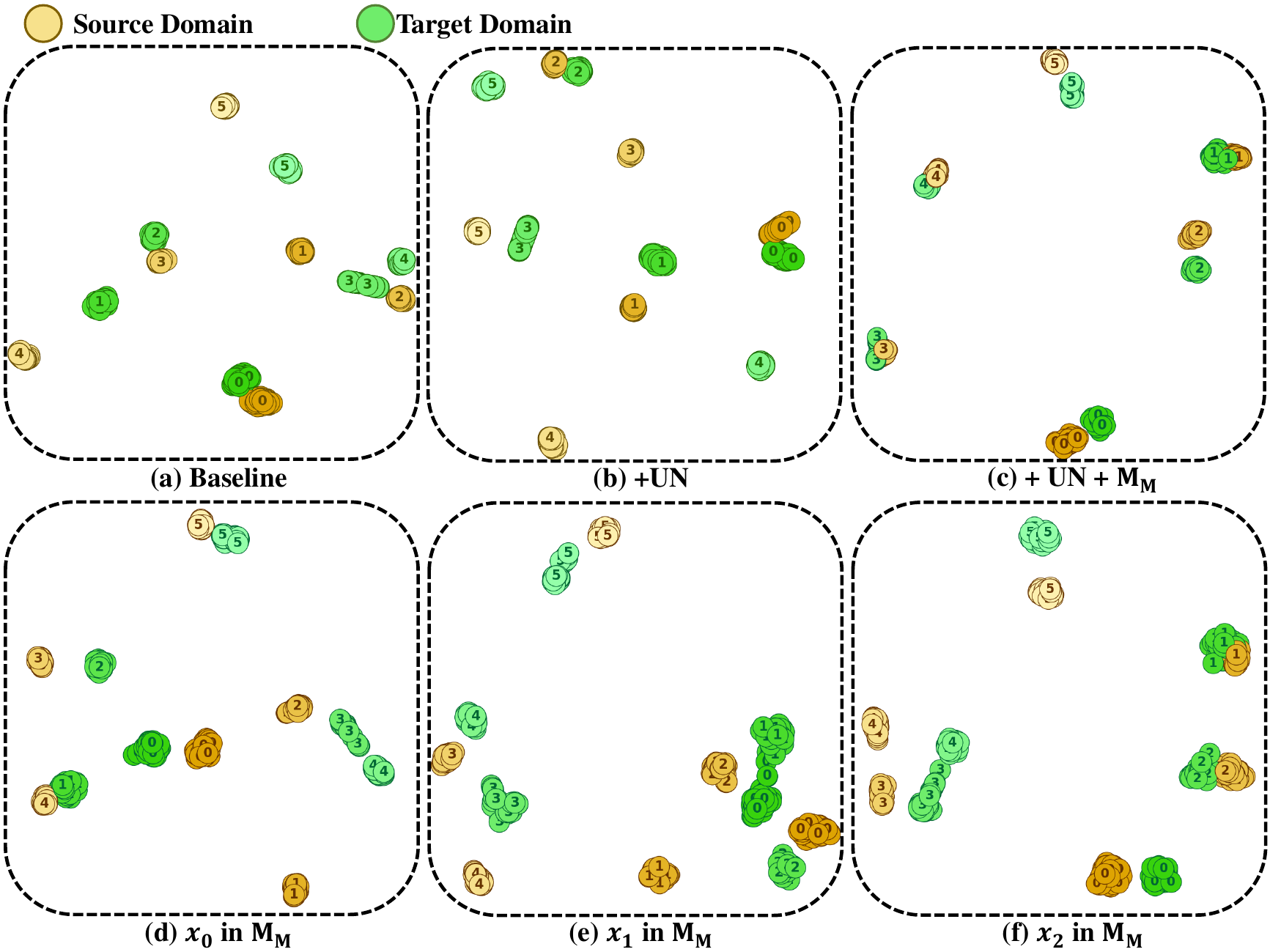}
    \caption{The figure illustrates the t-SNE visualization results of features extracted by different methods and across various domains on the UCIHAR dataset's 2\_to\_11 fold. The numbers in the markers represent class indexes. Panels (a) to (f) display features extracted by different methods: (a) uses the DDC method as the baseline, (b) adds uncertainty estimation, (c) adds uncertainty estimation and a multi-scale mixing architecture with max-pooling. Panels (d) to (f) show different features in different scales from (c).}
    \label{Figure:new_tsne}
\end{figure*}
 \begin{figure*}[ht]
  \centering
    \includegraphics[width=\textwidth]{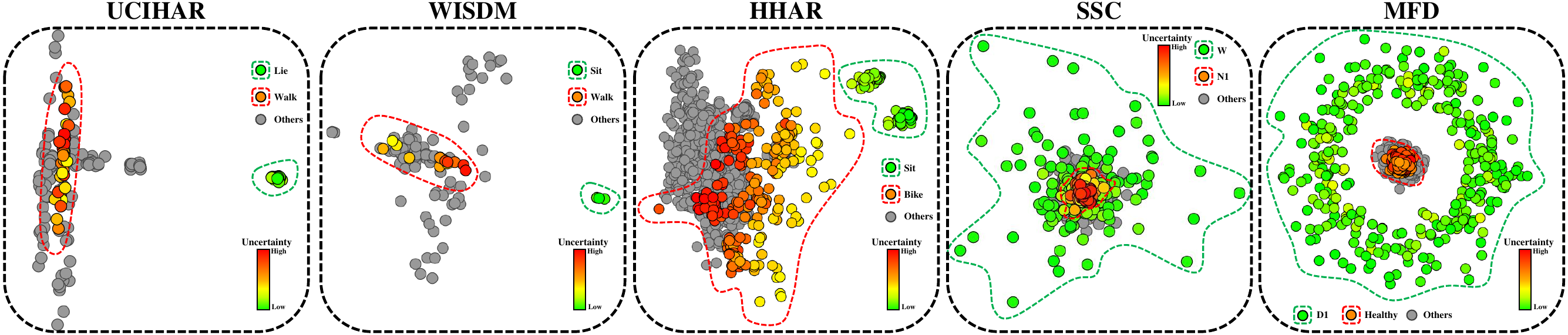}
    \caption{{The principal component analysis (PCA) visualization results of each sample and the uncertainty of some categories. The DDC~\cite{ddc} method was employed as a baseline for experiments on UCIHAR dataset's 6\_to\_23 fold, WISDM dataset's 33\_to\_12 fold, HHAR dataset's 4\_to\_6 fold, SSC dataset's 6\_to\_5 fold and MFD dataset's 0\_to\_3 fold.}}
    \label{Figure:pca}
\end{figure*}
The main contributions of this paper are summarized as follows:
\begin{itemize}
    \item We introduce a novel framework that incorporates uncertainty awareness for time series classification, enhancing predictive accuracy and reliability.
    
    \item We advocate for the use of mixed-length time series data as input, improving the model’s ability to learn from diverse patterns and cycles, thus enhancing the robustness of time series classification tasks.
    
    \item Our experimental evaluations across five distinct datasets—UCIHAR, WISDM, HHAR, SSC, and MFD—show that our method achieves a new state-of-the-art.
\end{itemize}

Parts of the results in this paper were previously presented in its workshop version, which received the Best Paper Award at the 4th International Workshop on Deep Learning for Human Activity Recognition, IJCAI 2024~\cite{liu2024uncertainty}. This paper significantly extends our earlier work in several crucial aspects:

\begin{itemize}
    \item We extend the proposed algorithm with using of mixed-length time series data encoded by Multi-scale Mixing Architecture as input, improving the model’s ability to learn from diverse patterns and cycles, thus enhancing the robustness of time series classification tasks. 

    \item We conduct a new experiment on new tasks such as sleep stage classification, and machine fault diagnosis to further evaluate our algorithm.

    \item We extend the investigation to new tasks including the Sleep Stage Classification (SSC)~\cite{eldele2022adast} and Machine Fault Diagnosis (MFD)~\cite{fd_dataset}. 
    \item Additionally, we have broadened our research framework by incorporating new analytical tasks. These enhancements are aimed at refining our experimental methods through the use of confusion matrices and model uncertainty assessments, as shown in Figures~\ref{Figure:matrix_uncertainty}, ~\ref{Figure:f1_uncertainty}, and ~\ref{Figure:uncertainty_density}. We also explore the shift in data distribution between the source and target datasets, illustrated in Figures~\ref{Figure:dist} and ~\ref{Figure:all_pca}. Furthermore, we conduct detailed performance comparisons (Figure~\ref{Figure:detail}), analyze Expected Calibration Error (ECE) in Figure~\ref{Figure:ece}, and use t-SNE to visualize features extracted by various methods (Figure~\ref{Figure:new_tsne}). Additionally, we assess the distribution of data through Principal Component Analysis (PCA) on model uncertainty, depicted in Figure~\ref{Figure:pca}. These approaches enhance the depth and accuracy of our evaluation, allowing for a more comprehensive understanding of the robustness of deep learning models in time series classification.

    \item We analyze the performance of the overall datasets and tasks, as shown in Tables~\ref{table:sota} and ~\ref{table:new_sota}, to provide a comprehensive evaluation of our algorithm.

\end{itemize}
\section{Related Work}
{
\subsection{Unsupervised Domain Adaptation}
Classical domain adaptation methods mitigate domain shift by aligning feature distributions between source and target domains. Techniques like CORAL~\cite{sun2016return} align second-order statistics (e.g., covariance matrices) to reduce linear discrepancies, while TCA~\cite{pan2010domain} projects data into a latent subspace using kernel methods to handle nonlinear shifts. Feature alignment methods minimize divergence metrics (e.g., MMD) or reweight samples to learn domain-invariant representations. However, these approaches often rely on shallow or linear transformations, limiting their ability to model complex, high-dimensional data. They may also require explicit domain labels, struggle with large domain gaps, and lack robustness to noisy target domains. Modern deep adaptation frameworks extend these ideas by leveraging deep networks to learn hierarchical, transferable features through adversarial training or deep alignment objectives.}

\subsection{Unsupervised Domain Adaptation for Time Series}
The time series classification tasks \cite{codats,cai2024fedcov,cai2024bayesian,liu2024harmonizing,qin2023cau,liu2020crnet} frequently exhibit significant domain shifts. This occurs when models trained on data from one subject (source domain) are applied to data from another (target domain).

While there have been significant achievements in computer vision, the adaptation of these successes to human activity recognition data remains limited. VRADA \cite{Purushotham2017VariationalRA} leverages a variational recurrent neural network (VRNN) and employs adversarial training to capture complex, domain-invariant temporal relationships. Building on VRADA, CoDATS \cite{Wilson2020MultiSourceDD} utilizes a convolutional neural network (CNN) as the feature extractor to enhance this methodology. Methods such as SASA \cite{Cai2021TimeSD} strive to align the conditional distribution of human activity recognition data by reducing the discrepancies in the associative structures of time series variables across domains. AdvSKM \cite{ijcai2021p378} and related works \cite{Ott2022DomainAF} employ metric-based approaches to align domains based on statistical divergence.
Techniques like DAF \cite{DAF-icml} extract both domain-invariant and domain-specific features, utilizing a shared attention module with a reconstruction task for forecasting across source and target domains. Contrastive DA methods such as CLUDA \cite{Ozyurt2022ContrastiveLF} and CLADA \cite{CALDA} employ augmentations to extract domain-invariant and contextual features for prediction. Furthermore, Maekawa \textit{et al.} \cite{maekawa2016toward} introduced an unsupervised method that identifies individual iterations of assembly work using acceleration data. However, the application of evidential learning to domain adaptation for human activity recognition remains unexplored. In our research, we apply evidential learning to the task of human activity recognition, with the objective of improving both the accuracy and reliability of the results.

\subsection{Uncertainty Estimation}
With the rapid development of deep learning~\cite{liu2020crnet,liu2024harmonizing,yue2024pyramid,liu2025modality}, estimating uncertainty in machine learning models has gained considerable attention, particularly in safety-critical domains such as autonomous vehicle navigation and medical diagnostics. This research explores uncertainty in two primary forms: aleatoric and epistemic. Aleatoric uncertainty, inherent in the observations like sensor noise, is contrasted with epistemic uncertainty, which pertains to model parameters and can decrease with additional data \cite{2018ensemble}.

Bayesian modeling highlights the significant memory and computational demands of traditional methods, often rendering them impractical. In response, some strategies involve training sub-networks with shared parameters, merging their outputs to address these limitations \cite{antoran2020depth}. Conversely, deterministic techniques provide direct uncertainty estimates, reducing model overconfidence \cite{van2020uncertainty}.

Murat et al.~\cite{2018Evidential} introduce a method to enhance prediction confidence by explicitly modeling prediction uncertainty through subjective logic and a Dirichlet distribution on class probabilities. This approach, distinct from conventional methods focused solely on minimizing prediction loss or Bayesian networks, shows enhanced performance in managing out-of-distribution queries and withstanding adversarial attacks.

In this paper, we propose employing these uncertainty estimation techniques for unsupervised domain adaptation in human activity recognition tasks, aiming to leverage these methodologies to improve model reliability and performance in critical applications.
\section{Methods}

\subsection{Task Definition}

In this paper, we delineate the concept of unsupervised domain adaptation (UDA). The scenario considered involves access to labeled data from a source domain  { $X^s = \{x_i^s, y_i^s\}_{i=1}^{N_s}$, encompassing either univariate or multivariate human activity recognition data, and unlabeled data from a target domain $X^s = \{x_j^t\}_{j=1}^{N_t}$.} Here, $N_s$ and $N_t$ signify the number of samples in the source and target domains, respectively. Our focus is on classification tasks, with the assumption that both domains share the same set of labels, denoted by $Y = \{1, 2, \ldots, K\}$, where $K$ represents the total number of classes.

Following data preprocessing methods~\cite{adatime}, we divide the source domain into a training subset $X_{tr}^s$ with $N_{tr}^s$ samples, and a test subset $X_{te}^s$ with $N_{te}^s$ samples. Similarly, the target domain is partitioned into a training subset $X_{tr}^t$ with $N_{tr}^t$ samples, and a test subset $X_{te}^t$ with $N_{te}^t$ samples. It is essential to note that the source and target domains are characterized by distinct marginal distributions, denoted as $P_s(x) \neq P_t(x)$, although the conditional distribution is assumed to remain consistent, i.e., $P_s(y|x) = P_t(y|x)$.

The primary objective of UDA is to minimize the discrepancy between the distributions $P_s(x)$ and $P_t(x)$, under the premise that the label spaces are identical across the domains. Denote the discrepancy between the marginal distributions $P_s(x)$ and $P_t(x)$ as $L_{d}$. One can typically use the different orders of data distribution statistics to calculate $L_{d}$.
A combined loss of source classification $L_{cls}$ and the domain alignment $L_{d}$ is jointly optimized to train a model $M=h(f(\theta))$ with a feature extractor $f(\theta)$ and a classifier $h(\theta)$ as typically shown in Figure~\ref{Figure:pipeline}. The combined loss $L_{uda}$ achieves a certain balance between the accuracy of the prediction in the source and the regularization of domain alignment:
\begin{equation}
\label{loss_UDA}
L_{uda} = \lambda_1L_{cls}+\lambda_2L_{d},
\end{equation}
where $\lambda_1$ and $\lambda_2$ are the regularization parameters.

\subsection{The Multi-scale Mixing Architecture}

Time series data exhibit unique characteristics at different scales; fine scales capture detailed patterns while coarse scales are indicative of broader, macroscopic variations~\cite{mozer1991induction}. This multi-scale perspective is crucial as it helps disentangle the intricate variations inherent in temporal data, which is particularly beneficial for modeling temporal variations in time series classification tasks.
We introduce the Multi-scale Mixing Architecture, a novel architecture designed for multi-scale time series analysis that optimally utilizes both past data extraction and future prediction, as shown in Figure~\ref{Figure:pipeline}. The initial step involves downsampling the entire time sequence, represented as $x \in \mathbb{R}^{P \times C}$, into multiple scales through average pooling, resulting in a set of multi-scale time series $X = \{x_0, \ldots, x_M\}$, where $x_m \in \mathbb{R}^{\lfloor \frac{P}{2^m} \rfloor \times C}$ for $m \in \{0, \ldots, M\}$, and $C$ denotes the number of variables. 

Subsequently, we individually encode and decode the multi-scale features $f_i$, then integrate these feature maps for the final prediction. The series $x_0 = x$ represents the finest temporal variations as the input series, while the highest-level series $x_M$ captures macroscopic variations. These multi-scale series are transformed into deep features, independently optimized, and then aggregated into final feature maps for prediction and are the main components for calculating the $L_{cls}$ after through a classification header. Additionally, auxiliary classification losses from features at different scales are introduced through auxiliary heads to support the training process. These auxiliary losses help guide the model in learning more effectively from multi-scale features. Together with the main classification component, the final classification loss, $L_{cls}$, is composed of multiple terms:
\[L_{cls} = L_{cls}^{Y^s}+\lambda_{i} \sum L_{cls}^{Y_i^s},\] where for $i\in\{0, 1, 2\}$, $\lambda_{i}$ takes the values $\{0.5, 0.25, 0.25\}$ in our experiments. { As illustrated in Fig. \ref{Figure:pipeline}, $L_{cls}^{Y^s}$ is the classification loss using the combined feature from all scales from the ``Final Header", while $L_{cls}^{Y_i^s}$ is the classification loss at different scale from ``Aux Headers".}  This method ensures a comprehensive representation of the input series across different scales, thereby optimizing both the fidelity of temporal pattern recognition and the predictive accuracy of the model.

In this paper, we explore different down-sampling strategies, such as CNN-only, max-pooling, random-pooling, and average-pooling,  in the multi-scale mixing architecture to handle time series inputs at various resolutions, as shown in Fig. \ref{Figure:pooling}, where each subfigure represents a unique combination of down-sampling techniques applied to multi-scale input.  It compares five approaches: (a) down-sampling with only 1D CNN layers, (b) combining 1D CNN with max-pooling, (c) using random-pooling after CNN, (d) applying max-pooling alone, and (e) utilizing average-pooling. Each method processes multi-scale inputs $(x_0, x_1, x_2)$ to capture features across varying temporal resolutions, highlighting how different pooling strategies affect feature extraction and down-sampling. Parameters of the different down-sampling methods in Fig. \ref{Figure:pooling} are summarized in Appendix for reproduction.

\subsection{Target Modeling with Uncertainty} 
In unsupervised domain adaptation (UDA), models are typically trained to align features across domains via adversarial training or distribution matching, as shown in Fig.~\ref{Figure:pipeline}. However, even well-aligned features can correspond to ambiguous or novel patterns, leading to overconfident mispredictions. We argue that UDA frameworks must quantify uncertainty explicitly to identify unreliable predictions caused by domain shifts to ensure calibrated outputs. Building on this principle, we enhance the UDA pipeline in Fig.~\ref{Figure:pipeline} with evidential learning. Traditional classifiers output softmax probabilities as point estimates, conflating confidence with certainty. Evidential learning instead treats predictions as accumulated evidence under a Dirichlet prior over class probabilities. This approach identifies misaligned or novel samples with high-uncertainty and prioritizes them to learning uncertainty-aware features, enabling cautious adaptation to ambiguous data while reducing overconfidence.

Following the approach of evidential learning for classification introduced in \cite{2018Evidential}, we model the labels with a multinomial/categorical distribution:
\begin{equation}
f({\bf p}|{\bf \alpha})=\frac{1}{ {B}(\alpha)}{\prod_{k=1}^{K}{p_k}^{\alpha_k-1}},
\end{equation}
where $B(\alpha)=\frac{\prod_{k=1}^{K}\Gamma(\alpha_k)}{\Gamma(\sum_{k=1}^{K}\alpha_k)}$ is the  K-dimensional multinomial beta function, $\Gamma(\cdot)$ is the Gamma function and ${\bf p} \in{\mathcal S}_K$ with ${\mathcal S}_K$ being the $K$-dimensional unit simplex defined as \[{\mathcal S}_K=\left\{  {\bf p} |{\sum_{k=1}^{K} p_k} = 1, with~ 0 \le p_1,\cdots, p_k \le 1\right\}.\]
The prior reflects our beliefs about the space of possible models and incorporates epistemic uncertainty in the problem formulation. This choice of Dirichlet prior due to its conjugate property with multinomial distribution, enables a close-form Dirichlet posterior and thus efficient inference.  The expected probability for the $k$-th outcome is therefore the mean of the Dirichlet posterior distribution:
\begin{equation}
\mathbb{E}\left[ p_k \right] = \hat{p}_k = \frac{\hat{\alpha}_k}{S},
\end{equation}
where $S=\sum_{k=1}^{K} \hat{\alpha}_k$ is the total amount of the evidence and $\hat{\alpha}_k$ is updated $\alpha_k$ from the posterior and $\mathbb{E}\left[\cdot\right]$ is the operation of expectation.

Both the Type-II Maximum Likelihood in Equation~(\ref{eq_typeII}) and two different Bayesian risks using cross-entropy error in Equation~(\ref{eq_ce}) and mean square error in Equation~(\ref{eq_mse}) can be used \cite{2018Evidential} to design the loss to estimate $\hat{\alpha}_k$ from the network $M$.

\noindent \textbf{Type-II Maximum Likelihood Loss:}
\begin{align} \label{eq_typeII}
L_i^{ml}(\theta) &= -log\left(  \int_{k=1}^{K}p_{ik}^{y_{ik}}\frac{1}{B(\boldsymbol \alpha_i)}\prod_{k=1}^{K}p_{ik}^{\alpha_{ik}}d{{\bf p}_i}\right) \nonumber \\& =\sum_{k=1}^{K}y_{ik}\left(  log(S_i)-log(\alpha_{ik})\right),
\end{align}
where $S_i$ is the total evidence associated with sample $i$ and $\boldsymbol \alpha_i$ is the vector of $K$ for sample $i$.

\noindent \textbf{Bayesian risks using cross-entropy error:}
\begin{align} \label{eq_ce}
L_i^{ce}(\theta)&=\int\left[ \sum_{k=1}^{K}-y_{ik}log(p_{ik}) \right]\frac{1}{B({\boldsymbol \alpha}_i)}\prod_{k=1}^{K}p_{ik}^{\alpha_{ik}-1}d{\bf p}_i \nonumber \\
&= \sum_{k=1}^{K}y_{ik}\left(\varphi(S_i)-\varphi(\alpha_{ik})  \right),
\end{align}
where $\varphi(\cdot)$ is  the digamma function.

\noindent \textbf{Bayesian risks using mean square error:}
\begin{align}\label{eq_mse}
L_i^{mse}(\theta)&=\int\left\| {\bf y}_i-{\bf p}_i \right\|_{2}^{2}\frac{1}{B({\boldsymbol \alpha}_i)}\prod_{i=1}^{K}p_{ij}^{\alpha_{ij}-1}d{\bf p}_i \nonumber \\ 
&=\sum_{j=1}^{K}{\mathbb{E}}\left[ y_{ij}^2-2y_{ij}p_{ij}+p_{ij}^2 \right] \nonumber \\ 
&=\sum_{j=1}^{K}y_{ij}^2-2y_{ij}{\mathbb E}[p_{ij}]+{\mathbb E}[p_{ij}^2] \nonumber \\
&= \sum_{k=1}^{K}\left( y_{ik}- \hat p_{ik}\right)^2+\frac{\hat p_{ik}\left( 1-\hat p_{ik} \right)}{S_i+1}
\end{align}
where $\hat p_{ik} =\frac{\alpha_{ik}}{S_i}$ is the prediction. The loss  $L_i^{mse}$ aims to minimize jointly the prediction error and the variance of the Dirichlet experiment generated by the neural net for each sample $i$. We tried all three different losses in our experiments and empirically found that the Bayesian risk with cross-entropy error when optimized yields the best performance for our case. Please refer to Table~\ref{table:sota} later in the paper for the performance comparison on difference losses.

Following \cite{2018Evidential}, a Kullback-Leibler (KL) divergence term is further used in the loss function to regularize the predictive distribution space by penalizing parameters $\alpha_k$ that do not contribute to data fit:
\begin{align}
& KL\left[ {\mathcal D}({\bf p}_i|\tilde{\boldsymbol \alpha}_i)|| {\mathcal D}({\bf p}_i|{\boldsymbol 1})\right] \nonumber \\
& = log\left( \frac{\Gamma\left( \sum_{k=1}^{K} \tilde{\alpha}_{ik}\right)} {\Gamma(K)\prod_{k=1}^{K}\Gamma(\tilde{\alpha}_{ik})} \right) \nonumber \\ & ~~~~~~~+\sum_{k=1}^{K}\left( \tilde{\alpha}_{ik}-1 \right)\left[ \varphi(\tilde{\alpha}_{ik})-\varphi\left(  \sum_{j=1}^{K}\tilde{\alpha}_{ij}\right) \right],
\end{align}
where ${\mathcal D}({\bf p}_i|{\boldsymbol 1})$ is the uniform Dirichlet distribution with $\boldsymbol 1$ being the $K$ ones vector, and  ${\tilde{\boldsymbol \alpha}}_{i} = {\bf y}_i +  \left( 1 - {\bf y}_i \right)\cdot {\boldsymbol \alpha}_{i}$ is the Dirichlet parameters after removal of the non-misleading evidence from predicted parameters ${\boldsymbol \alpha}_{i}$ for sample $i$.
Therefore the total evidential loss for $\alpha_k$ estimation is a balance between the Bayesian risk and the KL divergence:
\begin{align}\label{evi_loss}
L_{evi}=\sum_{i=1}^{N_s}L_i + \lambda_t\sum_{i=1}^{N_s}KL\left[ {\mathcal D}({\bf p}_i|\tilde{\boldsymbol \alpha}_i)|| {\mathcal D}({\bf p}_i|{\boldsymbol 1}) \right],
\end{align}
where $\lambda_t = min(1.0, t/10) \in [0, 1]$ is the annealing coefficient, $t$ is the index of the current training epoch.

Without introducing additional model parameters, the uncertainty for prediction is directly calculated by:
\begin{equation}\label{uncertainty}
u = \frac{K}{S_i} 
\end{equation}
according to the Subjective Logic framework, where ${S_i}=\sum_{k=1}^{K} \hat{\alpha}_{ik}$ is the total evidence of sample $i$. 
\subsection{Final Loss Function for UDA with Uncertainty}
The final loss of the proposed method combines the original loss of the UDA task in Equation~(\ref{loss_UDA}) and the evidential loss in Equation~(\ref{evi_loss}):
\begin{equation}\label{final_loss}
L = L_{uda} + \lambda_3 L_{evi} 
= \lambda_1L_{cls}+\lambda_2L_{d}+ \lambda_3 L_{evi},
\end{equation}
where $\lambda_1$, $\lambda_2$ and $\lambda_3$ balance the three loss components. In our experiments, $\lambda_1$ and $\lambda_2$ are directly set the same as the optimal ones given in \cite{ragab2023adatime} and optimal $\lambda_3$ is found empirically by the Target Risk (TGT) \cite{ragab2023adatime} by leaving out a subset of target domain samples and their labels as a validation set.
\section{Experiments}

\subsection{Datasets and Evaluation Metrics}

\noindent \textbf{UCIHAR:} The UCIHAR dataset~\cite{uciHAR_dataset} is collected using three types of sensors—an accelerometer, a gyroscope, and body sensors—attached to 30 subjects. Each subject performed six activities: walking, walking upstairs, walking downstairs, standing, sitting, and lying down. Due to the variability among subjects, each is treated as a separate domain. Following the approach in AdaTime~\cite{adatime}, we selected five scenarios for our study.

\noindent \textbf{WISDM:} Like UCIHAR, the WISDM dataset~\cite{wisdm_dataset} employs accelerometer sensors but includes 36 subjects performing the same activities. It introduces challenges such as class imbalances, where data from certain subjects may lack some activity classes. Following AdaTime~\cite{adatime}, we selected five cross-domain scenarios at random for analysis.

\noindent \textbf{HHAR:} The HHAR dataset~\cite{hhar_dataset}, or Heterogeneity Human Activity Recognition, includes data from nine subjects using both smartphone and smartwatch sensors. We followed AdaTime~\cite{adatime} in utilizing five randomly selected cross-domain scenarios for our experiments.

\noindent
\textbf{SSC:} The Sleep Stage Classification (SSC) involves categorizing EEG signals into five stages: Wake (W), three Non-Rapid Eye Movement (NREM) stages (N1, N2, N3), and Rapid Eye Movement (REM). We use the Sleep-EDF dataset~\cite{goldberger2000physiobank}, which contains EEG readings from 20 healthy subjects. We focused on a single EEG channel (Fpz-Cz) and included data from 10 subjects, setting up five cross-domain scenarios following prior research~\cite{eldele2021attention,adatime} for controlled comparisons across EEG stages and subjects.

\noindent 
\textbf{MFD:} The Machine Fault Diagnosis (MFD) dataset from Paderborn University~\cite{lessmeier2016condition} is used for detecting faults via vibration signals under four distinct operating conditions, each treated as a separate domain. Our analysis involves five cross-condition scenarios to evaluate domain adaptation effectiveness, with each sample containing 5120 data points from a univariate channel, in line with previous methodologies~\cite{adatime}.

\noindent \textbf{Analysis of data distribution:}
Data distribution analyses are given in Figure~\ref{Figure:dist} and Figure~\ref{Figure:all_pca}, where the label distributions and sample distributions after PCA for both source and target domains against different splits are visualized. It is clear that the UCIHAR and HHAR datasets have relatively uniform label distributions across various classes. In contrast, the other three datasets show more significant disparities among class label distributions. Additionally, the label distribution differences between the source and target domains for UCIHAR, HHAR, and MFD datasets are all less than 5\%, indicating minimal overall variation. While the distribution differences of various categories between source and target domains in the WISDM and SSC datasets are significant, which partially explains why the F1 score for these datasets, shown later in Table~\ref{table:new_sota}, are relatively low across different methods. The PCA visualization of sample distributions across different domain splits reveals substantial differences between source and target domains across multiple split configurations. This observation provides a strong motivation for domain adaptation strategies to align features across different domains, improving model robustness and generalization.

\noindent \textbf{Evaluation Metric:}
Following previous methods~\cite{ragab2023adatime}, to verify the effectiveness of the proposed framework with uncertainty awareness, we use 12 baseline models, including the base model without domain alignment, DDC\cite{ddc}, Deep\cite{sun2017correlation}, HoMM\cite{HoMM}, DANN\cite{DANN}, MMDA\cite{MMDA}, DSAN\cite{dsan}, CDAN\cite{CDAN}, DIRT\cite{shu2018dirt}, CoDATS\cite{codats}, AdvSKM\cite{advskm}, SASA\cite{sasa} and CLUDA~\cite{ozyurt2022contrastive}. The F1 score is reported as the evaluation metric to validate the performance.

\subsection{The Impact of Uncertainty}

\subsubsection{Impact on Prediction Accuracy} 
The F1 scores of different baselines incorporated with uncertainty estimation are given in Table~\ref{table:sota}, where uncertainty imposed by normal-inverse-gamma (NIG) prior \cite{deep_evidential} is used as a comparison. We implemented all three different evidential losses aforementioned in Equation \ref{eq_typeII}, Equation \ref{eq_ce}, and Equation \ref{eq_mse}, respectively.  The Bayesian risk loss with
cross-entropy shows the best F1 score on average.
All baseline models when incorporated with uncertainty outperform their counterparts. Baseline models with Dirichlet prior (DIR) outperform those with NIG prior, possibly because DIR is better suited for classification tasks. A significant performance improvement is observed even when there is no domain alignment, which suggests that the awareness of uncertainty itself imposes the model ability of generalization and robustness for domain shift.

The detailed performance comparisons with the proposed method and the baseline model without domain adaptation are summarized in Figure~\ref{Figure:matrix_uncertainty} and Figure~\ref{Figure:detail}, where the averaged classification confusion matrix over all folds and the F1 score over different folds are given, respectively. Higher positive prediction rates are achieved while miss-classification rates are decreased as shown in Figure~\ref{Figure:matrix_uncertainty}, especially for walk and upstairs labels in the UCIHAR dataset,  downstairs and upstairs labels in the WISDM dataset, and bike and upstairs labels in the HHAR dataset. The corresponding uncertainty values of each label average over the samples and folds are shown in the third column, where a higher uncertainty value is observed for a prediction with a low F1 score. Figure~\ref{Figure:detail} shows that our method consistently achieves higher F1 scores over all different folds. 

\subsubsection{Impact on Feature Learning}

The distribution of learned features of baseline method DDC and our method are compared in Figure~\ref{Figure:new_tsne}. In comparison to the baseline method, the inclusion of uncertainty awareness in our approach leads to a more concentrated aggregation of features belonging to the same category across different domains, which is particularly noticeable for categories 1 and 2. Additionally, it pushes apart the features from different categories, both within and between the source and target domains. This separation is especially evident for categories 3 and 4 in the target domain, and categories 2 and 3 between the source and target domains.

\subsection{The Impact of Multi-scale Mixing Architecture}

Table~\ref{table:new_sota} demonstrates that incorporating a multi-scale mixing architecture alongside uncertainty prediction significantly improves the average F1 scores of various baseline methods across different datasets. The results indicate that different down-sampling methods show varied performance across different datasets, the overall differences are minimal.  Figure~\ref{Figure:new_tsne} (c) further shows that when the multi-scale mixing architecture is integrated, this aggregation of features from multiple domains is further improved. Moreover, within the multi-scale architecture, higher-level sequences demonstrate superior category aggregation, indicating the effectiveness of our approach in enhancing feature learning across scales as shown from Figure~\ref{Figure:new_tsne} (d) to (f).

To quantitatively illustrate the impact of multi-scale mixing on domain shift, we employ domain discrepancy metrics such as Maximum Mean Discrepancy (MMD) and Wasserstein Distance (WD) to measure the distance between the source and target domains when applying multi-scale mixing. The results, presented in the Tables \ref{table:mmd} and \ref{table:wd}, respectively, the tables demonstrate that multi-scale mixing effectively reduces domain discrepancy, as quantified by both WD and MMD. Specifically, multi-scale mixing with 1D CNNs shows notable success in minimizing MMD, whereas implementations using max or average pooling are particularly effective at reducing WD.

\begin{table}[h]
\Huge
\centering
\caption{Effect of Multi-Scale Mixing Maximum Mean Discrepancy (MMD), the baseline method is DDC. While + $M_{L}$ denotes the multi-scale mixing architecture with learnable parameter (1D CNN), + $M_{R}$ denotes the multi-scale mixing architecture with random pooling, + $M_{M}$ denotes the multi-scale mixing architecture with max pooling, + $M_{A}$ denotes the multi-scale mixing architecture with averaging pooling}
\renewcommand\arraystretch{1} 
\resizebox{1.03\linewidth}{!}{
\begin{tabular}{l|cccccccccc|c}
\toprule
\multicolumn{12}{c}{\textbf{UCIHAR}} \\
\hline
Split&2to11&6to23&7to13&9to18&12to16&18to27&20to5&24to8&28to27&30to20&Mean\\
\hline
Baseline    &0.236 &0.259 &0.252 &0.219 &0.302 &0.234 &0.226 &0.288 &0.244 &0.262 &0.252 \\
$+M_L$      &0.182 &\textbf{0.160} &0.191 &\textbf{0.201} &\textbf{0.190} &\textbf{0.165} &\textbf{0.153} &0.226 &\textbf{0.198} &\textbf{0.169} &\textbf{0.184}\\
$+M_L+M_M$  &\textbf{0.162} &0.161 &0.231 &0.276 &0.203 &0.211 &0.180 &\textbf{0.209} &0.221 &0.170 &0.202 \\
$+M_R$      &0.238 &0.192 &0.219 &0.254 &0.231 &0.281 &0.204 &0.244 &0.220 &0.209 &0.229 \\
$+M_M$      &0.215 &0.217 &0.199 &0.248 &0.255 &0.254 &0.194 &0.239 &0.243 &0.203 &0.227 \\
$+M_A$      &0.223 &0.217 &\textbf{0.171} &0.263 &0.268 &0.250 &0.216 &0.223 &0.205 &0.222 &0.226 \\

\hline
\multicolumn{12}{c}{\textbf{WISDM}} \\
\hline
Split&7to18&20to30&35to31&17to23&6to19&2to11&33to12&5to26&28to4&23to32&Mean\\
\hline
Baseline    &0.213 &0.243 &0.249 &0.287 &0.268 &0.241 &0.269 &0.217 &0.243 &0.225 &0.245 \\
$+M_L$      &\textbf{0.144} &0.182 &0.194 &\textbf{0.180} &0.241 &\textbf{0.141} &0.243 &0.195 &\textbf{0.141} &0.199 &\textbf{0.186} \\
$+M_L+M_M$  &0.164 &\textbf{0.163} &0.243 &0.222 &0.277 &0.233 &\textbf{0.215} &\textbf{0.180} &0.177 &\textbf{0.196} &0.207\\ 
$+M_R$      &0.205 &0.226 &0.213 &0.263 &\textbf{0.233} &0.237 &0.330 &0.214 &0.200 &0.275 &0.240 \\
$+M_M$      &0.194 &0.215 &0.215 &0.248 &0.271 &0.197 &0.281 &0.207 &0.212 &0.261 &0.230 \\
$+M_A$      &0.221 &0.223 &\textbf{0.184} &0.241 &0.266 &0.214 &0.281 &\textbf{0.180} &0.179 &0.261 &0.225 \\

\hline
\multicolumn{12}{c}{\textbf{HHAR}} \\
\hline
Split&0to6&1to6&2to7&3to8&4to5&5to0&6to1&7to4&8to3&0to2&Mean\\
\hline
Baseline    &0.185 &0.183 &0.174 &0.194 &0.214 &0.206 &0.172 &0.230 &0.204 &0.191 &0.195 \\
$+M_L$      &\textbf{0.113} &\textbf{0.097} &\textbf{0.088} &\textbf{0.120} &0.191 &0.212 &\textbf{0.075} &\textbf{0.171} &\textbf{0.133} &0.176 &\textbf{0.138} \\
$+M_L+M_M$  &0.202 &0.120 &0.157 &0.131 &0.213 &\textbf{0.136} &0.157 &0.189 &0.187 &\textbf{0.146} &0.164\\ 
$+M_R$      &0.220 &0.106 &0.132 &0.126 &0.178 &0.213 &0.160 &0.224 &0.181 &0.147 &0.169 \\
$+M_M$      &0.188 &0.143 &0.133 &0.161 &0.184 &0.185 &0.153 &0.204 &0.149 &0.185 &0.168 \\
$+M_A$      &0.182 &0.161 &0.095 &0.179 &\textbf{0.156} &0.166 &0.138 &0.204 &0.162 &0.155 &0.160 \\

\hline
\multicolumn{12}{c}{\textbf{SSC}} \\
\hline
Split&0to11&7to18&9to14&12to5&16to1&3to19&18to12&13to17&5to15&6to2&Mean\\
\hline
Baseline    &0.185 &0.109 &0.093 &0.114 &0.116 &0.107 &0.118 &\textbf{0.162} &0.144 &0.112 &0.126 \\
$+M_L$      &\textbf{0.120} &\textbf{0.068} &0.062 &0.087 &0.114 &0.084 &0.062 &0.200 &\textbf{0.052} &0.091 &\textbf{0.094} \\
$+M_L+M_M$  &0.200 &0.109 &0.061 &\textbf{0.082} &0.107 &\textbf{0.055} &\textbf{0.057} &0.203 &0.118 &0.126 &0.112\\ 
$+M_R$      &0.181 &0.110 &0.065 &0.128 &0.129 &0.133 &0.124 &0.169 &0.107 &\textbf{0.063} &0.121 \\
$+M_M$      &0.186 &0.083 &0.075 &0.084 &\textbf{0.093} &0.112 &0.118 &0.187 &0.095 &0.103 &0.114 \\
$+M_A$      &0.153 &0.104 &\textbf{0.051} &0.088 &0.099 &0.122 &0.118 &0.187 &0.105 &0.070 &0.110 \\

\hline
\multicolumn{12}{c}{\textbf{MFD}} \\
\hline
Split&0to1&0to3&1to0&1to2&1to3&2to1&2to3&3to0&3to1&3to2&Mean\\
\hline
Baseline    &0.221 &0.223 &0.247 &0.257 &0.124 &0.196 &0.183 &0.224 &0.140 &0.254 &0.207 \\
$+M_L+M_M$  &0.183 &0.184 &\textbf{0.182} &0.219 &0.086 &0.135 &\textbf{0.136} &\textbf{0.172} &\textbf{0.057} &0.205 &0.156\\ 
$+M_R$      &0.204 &0.208 &0.198 &\textbf{0.150} &0.071 &0.168 &0.173 &0.215 &0.064 &\textbf{0.160} &0.161 \\
$+M_M$      &\textbf{0.165} &\textbf{0.169} &0.192 &0.179 &\textbf{0.061} &0.144 &\textbf{0.136} &0.184 &0.068 &0.179 &\textbf{0.148} \\
$+M_A$      &0.173 &0.193 &0.198 &0.190 &0.079 &\textbf{0.128} &0.139 &0.206 &0.060 &0.208 &0.157 \\

\bottomrule
\end{tabular}
}
\label{table:mmd}
\end{table}
\begin{table}[h]
\Huge
\centering
\caption{Effect of Multi-Scale Mixing on Wasserstein Distance (WD), the baseline method is DDC. While + $M_{L}$ denotes the multi-scale mixing architecture with learnable parameter (1D CNN), + $M_{R}$ denotes the multi-scale mixing architecture with random pooling, + $M_{M}$ denotes the multi-scale mixing architecture with max pooling, + $M_{A}$ denotes the multi-scale mixing architecture with averaging pooling}
\renewcommand\arraystretch{1} 
\resizebox{1.03\linewidth}{!}{
\begin{tabular}{l|cccccccccc|c}
\toprule
\multicolumn{12}{c}{\textbf{UCIHAR}} \\
\hline
Method&2to11&6to23&7to13&9to18&12to16&18to27&20to5&24to8&28to27&30to20&Mean\\
\hline
Baseline    &11.74&\textbf{15.39}&14.74&23.30&34.00&21.10&21.37&\textbf{18.22}&9.44&27.59&19.69 \\
$+M_L$      &\textbf{7.48}&20.36&\textbf{14.61}&21.56&31.59&\textbf{13.35}&17.36&21.02&18.19&\textbf{21.98}&18.75 \\
$+M_L+M_M$  &7.62&19.79&15.28&21.56&31.80&15.43&17.68&21.49&17.91&22.86&19.14 \\
$+M_R$      &8.23&19.67&14.99&21.47&\textbf{31.33}&14.56&18.20&20.47&16.81&22.50&18.82 \\
$+M_M$      &7.81&19.34&14.98&21.03&31.68&13.71&17.41&20.74&16.90&22.06&\textbf{18.57} \\
$+M_A$      &7.84&20.01&15.31&\textbf{20.71}&32.06&13.67&\textbf{17.31}&20.98&\textbf{16.29}&22.48&18.66 \\

\hline
\multicolumn{12}{c}{\textbf{WISDM}} \\
\hline
Method&7to18&20to30&35to31&17to23&6to19&2to11&33to12&5to26&28to4&23to32&Mean\\
\hline
Baseline    &30.45&34.58&30.27&41.04&\textbf{35.13}&27.95&38.71&21.61&\textbf{29.35}&49.59&33.87 \\
$+M_L$      &27.26&29.57&25.00&\textbf{21.45}&37.79&27.60&\textbf{28.93}&14.40&31.10&34.22&27.73 \\
$+M_L+M_M$  &27.49&29.46&25.57&23.80&37.78&28.83&30.09&15.45&33.61&34.13&28.62 \\
$+M_R$      &25.95&30.55&25.29&22.72&36.33&\textbf{27.41}&29.34&15.22&31.66&33.32&27.78 \\
$+M_M$      &\textbf{25.93}&29.97&25.75&22.15&36.45&27.79&28.95&14.93&31.97&33.14&27.70 \\
$+M_A$     &26.08&\textbf{29.40}&\textbf{24.92}&22.82&36.91&27.44&29.38&\textbf{14.16}&31.27&\textbf{33.13}&\textbf{27.55} \\

\hline
\multicolumn{12}{c}{\textbf{HHAR}} \\
\hline
Method&0to6&1to6&2to7&3to8&4to5&5to0&6to1&7to4&8to3&0to2&Mean\\
\hline
Baseline    &58.10&36.35&48.10&31.87&37.71&82.69&32.58&43.09&39.58&44.80&45.49 \\
$+M_L$      &29.77&\textbf{27.19}&29.41&18.17&21.59&42.32&22.44&25.96&24.98&28.23&27.01 \\
$+M_L+M_M$  &32.77&31.20&31.71&18.36&25.28&45.02&\textbf{21.33}&24.35&25.59&30.39&28.60 \\
$+M_R$      &28.68&29.48&28.51&17.08&\textbf{19.42}&\textbf{40.77}&23.51&23.12&24.47&29.33&26.44 \\ 
$+M_M$      &28.01&28.31&28.18&17.05&20.51&41.34&22.39&22.64&\textbf{23.68}&\textbf{27.74}&25.98 \\
$+M_A$      &\textbf{27.66}&27.53&\textbf{27.88}&\textbf{16.48}&20.24&41.19&22.96&\textbf{22.16}&24.26&28.22&\textbf{25.86} \\

\hline
\multicolumn{12}{c}{\textbf{SSC}} \\
\hline
Method&0to11&7to18&9to14&12to5&16to1&3to19&18to12&13to17&5to15&6to2&Mean\\
\hline
Baseline    &13.02&5.37&4.04&5.76&4.22&\textbf{4.73}&\textbf{6.30}&7.17&10.37&4.50&6.55 \\
$+M_L$      &\textbf{9.38}&\textbf{5.13}&3.02&4.04&4.22&7.07&8.47&7.45&8.15&5.40&6.23 \\
$+M_L+M_M$  &9.40&5.74&3.65&4.33&4.61&6.79&9.24&7.02&7.83&5.91&6.45 \\
$+M_R$      &9.41&5.24&2.71&3.71&4.46&7.34&8.90&7.21&8.36&5.37&6.27 \\
$+M_M$      &9.54&5.41&2.92&4.14&4.44&6.62&8.72&7.04&7.86&5.47&6.22 \\
$+M_A$      &9.84&5.88&\textbf{2.30}&\textbf{3.65}&\textbf{4.01}&7.01&8.78&\textbf{6.89}&\textbf{7.65}&\textbf{5.34}&\textbf{6.14} \\

\hline
\multicolumn{12}{c}{\textbf{MFD}} \\
\hline
Method&0to1&0to3&1to0&1to2&1to3&2to1&2to3&3to0&3to1&3to2&Mean\\
\hline
Baseline    &17.82&17.04&21.79&14.27&1.21&9.39&8.10&19.24&1.75&16.27&12.69 \\
$+M_L$      &\textbf{8.35}&7.88&11.44&8.53&4.55&4.95&\textbf{2.02}&10.69&2.12&5.76&6.63 \\
$+M_L+M_M$  &12.71&7.30&9.67&8.99&2.30&6.21&5.47&9.81&1.78&7.00&7.12 \\
$+M_R$      &8.97&\textbf{4.06}&9.07&5.88&0.98&3.87&2.57&8.34&1.54&5.18&5.05 \\
$+M_M$      &9.54&4.57&8.97&\textbf{5.72}&1.02&\textbf{3.37}&2.39&7.76&\textbf{0.97}&4.83&4.91 \\
$+M_A$      &9.43&4.30&\textbf{8.48}&6.02&\textbf{0.63}&3.76&2.86&\textbf{7.18}&1.37&\textbf{4.49}&\textbf{4.85} \\

\bottomrule
\end{tabular}
}
\label{table:wd}
\end{table}

\subsection{Model Performance Analysis}
To demonstrate the calibration of our model compared to the baseline model, we examine two key perspectives across various datasets in the target domain. Firstly, we empirically analyze the relationship between uncertainty estimates and prediction F1 scores. Secondly, we evaluate the Expected Calibration Error (ECE) metric.

\subsubsection{Uncertainty vs F1}

Our experiment results, as depicted in Figure~\ref{Figure:f1_uncertainty}, consistently demonstrate that uncertainty values exhibit a roughly linear inverse relationship with the F1 score across all five datasets. This suggests a broader phenomenon rather than being specific to particular data types. Lower uncertainty values correspond to higher confidence in model predictions. Conversely, when the model tends to make less accurate predictions, resulting in lower F1 scores, our model produces higher uncertainty values. This indicates that the uncertainty is well-calibrated.
\subsubsection{Expected Calibration Error}

The Expected Calibration Error (ECE) metric measures the discrepancy between the predicted probabilities and the true probabilities of the model. It provides a holistic assessment of the model's calibration performance across different confidence levels. The ECE metrics for different datasets are given in Figure~\ref{Figure:ece}.

\subsection{Uncertainty Density between Domains}
The byproduct of our method is to quantify the uncertainty of the prediction. Illustrations of the uncertainty density of the prediction on the source and target domain of different datasets are given in Figure~\ref{Figure:uncertainty_density}. The distributions of prediction uncertainty for the source and target domains are located in different ranges, with higher uncertainty observed in the predictions of the target domain compared to the source domain. This observation is expected and suggests that domain shift is still there even after domain alignment and interpretation of the predictions should be used cautiously, especially for those with higher uncertainty.

\subsection{Data distribution vs Uncertainty}
Figure~\ref{Figure:pca} visualizes various sample categories using principal component analysis (PCA), illustrating the relationship between data distribution and uncertainty. The color scale, ranging from green (low uncertainty) to red (high uncertainty), indicates the level of uncertainty associated with each sample. Higher uncertainty (redder colors) suggests that the model finds these samples more difficult to classify accurately, likely due to overlap or ambiguity in feature space. For example, in HHAR dataset, samples in ``Sit" represented in green, are located in distinct clusters with low uncertainty (indicated by the green color), suggesting that ``Sit" samples are well-separated from the other categories and are easier for the model to classify with confidence. While samples ``Bike" shown in orange, are overlapping with samples from other categories with moderate to high uncertainty. This suggests that ``Bike" samples are relatively more difficult to classify.
\section{Conclusion}
In this paper, we propose evidential learning in the framework of unsupervised domain adaptation for human activity recognition tasks. The uncertainty is incorporated using the network output via evidential loss without any additional network parameters to optimize the Bayesian risk under the Dirichlet prior to labels. The awareness of the uncertainty in the model prediction helps to learn features that are better aligned for source and target domains and thus show better generation ability in the target domain. 
Furthermore, we introduce a multi-scale mixing architecture to further enhance the model's performance. Extensive experiments conducted on 5 datasets and 12 baseline methods demonstrate that our proposed method enhances the robustness of the model in domain shift scenarios. Extensive analysis was conducted on the impact of uncertainty on feature distribution, revealing insights into how uncertainty influences feature alignment across different domains and enhances model robustness to domain shifts. Additionally, our analysis of the relationship between sample uncertainty and PCA of sample distributions indicates a strong association between uncertainty and sample difficulty. The PCA results show that higher uncertainty often corresponds to more challenging samples, suggesting a meaningful link between uncertainty levels and sample complexity.

\section*{Acknowledgment}
This work was supported by the Natural Science Foundation of Jiangxi Province (Grant No. 20223AEI91002). It was also partially supported by the National Natural Science Foundation of China (Grant No. 62461028) and the China Postdoctoral Science Foundation (Grant No. 2024T170364).
\section*{Appendix} \label{sec: Appendix_A}

\subsection{Comparison of Monte Carlo Uncertainty}
\begin{figure*}[ht]
  \centering
    \includegraphics[width=0.95\linewidth]{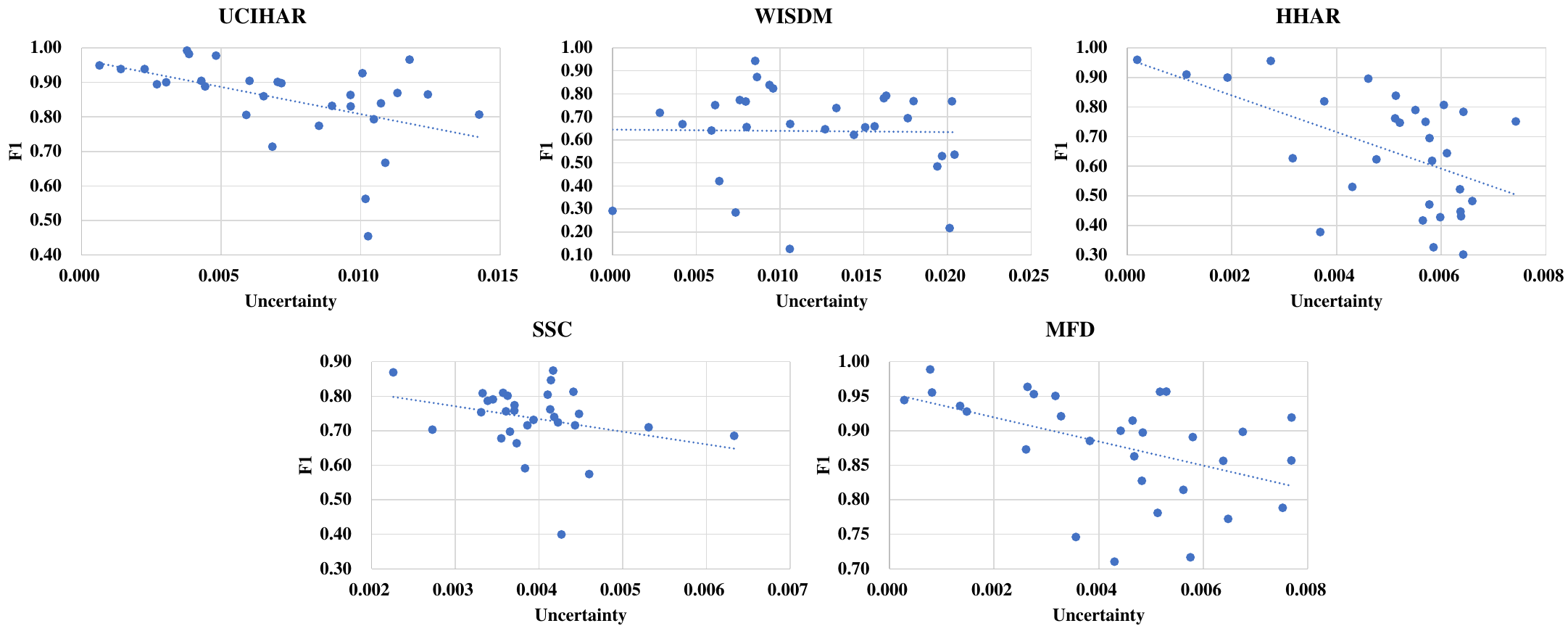}
    \caption{The correlation between F1 score and Monte Carlo uncertainty. The baseline is the basic model without domain adaption, randomly select 30 folds on UCIHAR, WISDM, HHAR, SSC and MFD datasets for test, each fold test 100 times. The x axis is the uncertainty obtained by MC dropout, and the y axis is the mean of the F1 score.}
    \label{Figure:MonteCarlo}
\end{figure*}

In addition to our evidential learning-based uncertainty, we also study uncertainty estimated using Monte Carlo (MC) dropout shown in Figure~\ref{Figure:MonteCarlo}. The results demonstrate that our method provides more interpretable uncertainty, exhibiting a stronger correlation with the F1 score compared to the MC approach.

\subsection{The Notational Summary Table}

We have show the Notational Summary in Table~\ref{Table:Notational}. 

\begin{table}[h]
\centering

\caption{Notational summary of the unsupervised domain adaptation framework}

\begin{tabular}{|c|l|}
\hline
\textbf{Symbol} & \textbf{Description}                       \\ \hline
$X_s$           & Source domain data                         \\ \hline
$Y_s$           & Labels for source domain data              \\ \hline
$X_t$           & Target domain data (unlabeled)             \\ \hline
$\mathcal{L}_d$ & Domain loss                                \\ \hline
$\mathcal{L}_{cls}$ & Classification loss on the source domain \\ \hline
$\mathcal{L}_{levi}$ & Evidential loss for uncertainty estimation \\ \hline
$f(\theta)$     & Feature extractor                          \\ \hline
$h(\theta)$     & Classifier                                 \\ \hline
$\alpha$        & Dirichlet distribution parameters          \\ \hline
$K$             & Total number of classes                    \\ \hline
$N_s, N_t$      & Number of samples in source and target domains \\ \hline
\end{tabular}

\label{Table:Notational}
\end{table}

\begin{table*}[h]
\Huge
\centering
\caption{Class distribution of different training and test sets in the UCIHAR dataset. The correspondence between class indices and class names is as follows: 0-`walk', 1-`upstairs', 2-`downstairs', 3-`sit', 4-`stand', 5-`lie'.}
\renewcommand\arraystretch{1} 
\resizebox{0.7\linewidth}{!}{
\begin{tabular}{c|cccccc|c|c|cccccc|c}
\toprule
\multicolumn{16}{c}{\textbf{UCIHAR}} \\
\hline
\multirow{2}{*}{\textbf{Train}}&\multicolumn{6}{c|}{\textbf{Class Indices}}&\multirow{2}{*}{\textbf{Sum}}&\multirow{2}{*}{\textbf{Test}}&\multicolumn{6}{c|}{\textbf{Class Indices}}&\multirow{2}{*}{\textbf{Sum}}\\
\cline{2-7}\cline{10-15}
&0&1&2&3&4&5&&&0&1&2&3&4&5&\\
\hline
Split\_1&66&37&34&33&37&35&242&Split\_1&29&16&15&14&16&15&105\\
Split\_2&41&34&33&32&38&33&211&Split\_2&18&14&14&14&16&15&91\\
Split\_3&41&41&34&36&43&43&238&Split\_3&17&18&15&16&18&19&103\\
Split\_4&42&36&31&35&39&38&221&Split\_4&18&16&14&15&17&16&96\\
Split\_5&39&33&33&31&39&36&211&Split\_5&17&14&14&13&17&16&91\\
Split\_6&40&36&33&38&40&40&227&Split\_6&17&15&15&17&17&17&98\\
Split\_7&40&36&33&33&37&36&215&Split\_7&17&15&14&15&16&16&93\\
Split\_8&33&29&26&32&38&38&196&Split\_8&15&12&12&14&16&16&85\\
Split\_9&36&34&29&35&32&35&201&Split\_9&16&15&13&15&13&15&87\\
Split\_10&37&33&26&38&31&40&205&Split\_10&16&14&12&16&13&18&89\\
Split\_11&41&38&32&37&33&40&221&Split\_11&18&16&14&16&14&17&95\\
Split\_12&35&36&32&36&43&42&224&Split\_12&15&16&14&15&18&18&96\\
Split\_13&40&38&33&34&40&43&228&Split\_13&17&17&14&15&17&19&99\\
Split\_14&41&38&31&38&42&36&226&Split\_14&18&16&14&16&18&15&97\\
Split\_15&38&34&29&41&37&50&229&Split\_15&16&14&13&18&16&22&99\\
Split\_16&36&36&33&48&54&49&256&Split\_16&15&15&14&21&24&21&110\\
Split\_17&43&33&32&45&54&50&257&Split\_17&18&15&14&19&24&21&111\\
Split\_18&39&41&38&40&51&45&254&Split\_18&17&17&17&17&22&20&110\\
Split\_19&37&28&27&51&51&58&252&Split\_19&15&12&12&22&22&25&108\\
Split\_20&36&36&31&46&51&47&247&Split\_20&15&15&14&20&22&21&107\\
Split\_21&36&33&32&59&62&63&285&Split\_21&16&14&13&26&27&27&123\\
Split\_22&32&30&25&43&44&50&224&Split\_22&14&12&11&19&19&22&97\\
Split\_23&41&36&38&48&47&50&260&Split\_23&18&15&16&20&21&22&112\\
Split\_24&41&41&38&48&48&50&266&Split\_24&17&18&17&20&21&22&115\\
Split\_25&52&45&41&45&52&51&286&Split\_25&22&20&17&20&22&22&123\\
Split\_26&41&38&35&55&52&53&274&Split\_26&18&17&15&23&22&23&118\\
Split\_27&40&35&31&49&56&52&263&Split\_27&17&16&13&21&24&22&113\\
Split\_28&38&36&32&50&55&56&267&Split\_28&16&15&14&22&24&24&115\\
Split\_29&37&34&34&42&45&48&240&Split\_29&16&15&14&18&20&21&104\\
Split\_30&46&46&43&43&41&49&268&Split\_30&19&19&19&19&18&21&115\\

\bottomrule
\end{tabular}
}
\label{table:ucihar-data}
\end{table*}  
\begin{table*}[h]
\Huge
\centering
\caption{Class distribution of different training and test sets in the WISDM dataset. The correspondence between class indices and class names is as follows: 0-`walk', 1-`jog', 2-`sit', 3-`stand', 4-`upstairs', 5-`downstairs'.}
\renewcommand\arraystretch{1} 
\resizebox{0.7\linewidth}{!}{
\begin{tabular}{c|cccccc|c|c|cccccc|c}
\toprule
\multicolumn{16}{c}{\textbf{WISDM}} \\
\hline
\multirow{2}{*}{\textbf{Train}}&\multicolumn{6}{c|}{\textbf{Class Indices}}&\multirow{2}{*}{\textbf{Sum}}&\multirow{2}{*}{\textbf{Test}}&\multicolumn{6}{c|}{\textbf{Class Indices}}&\multirow{2}{*}{\textbf{Sum}}\\
\cline{2-7}\cline{10-15}
&0&1&2&3&4&5&&&0&1&2&3&4&5&\\
\hline
Split\_0&12&48&0&0&13&56&129&Split\_0&7&26&0&0&7&30&70\\
Split\_1&0&51&0&0&0&51&102&Split\_1&0&28&0&0&0&27&55\\    
Split\_2&15&48&6&12&14&57&152&Split\_2&7&26&4&7&8&30&82\\
Split\_3&7&3&5&0&6&26&47&Split\_3&4&2&3&0&3&14&26\\
Split\_4&14&28&7&7&14&53&123&Split\_4&7&15&4&3&8&29&66\\
Split\_5&7&51&7&3&7&53&128&Split\_5&3&28&4&1&4&29&69\\
Split\_6&10&40&10&10&15&48&133&Split\_6&5&21&6&5&9&26&72\\
Split\_7&14&45&12&14&19&74&178&Split\_7&8&24&6&8&10&40&96\\
Split\_8&0&0&0&0&0&56&56&Split\_8&0&0&0&0&0&30&30\\
Split\_9&16&53&0&6&18&57&150&Split\_9&9&28&0&4&10&30&81\\
Split\_10&11&54&0&0&19&52&136&Split\_10&6&29&0&0&10&29&74\\
Split\_11&12&53&10&7&11&47&140&Split\_11&7&29&5&4&6&25&76\\
Split\_12&18&54&5&6&20&57&160&Split\_12&10&29&3&4&11&30&87\\
Split\_13&12&57&0&0&35&61&165&Split\_13&7&31&0&0&19&32&89\\
Split\_14&7&56&0&0&9&50&122&Split\_14&4&30&0&0&5&27&66\\
Split\_15&7&0&13&8&6&54&88&Split\_15&4&0&7&5&3&29&48\\
Split\_16&16&12&0&0&25&42&95&Split\_16&9&7&0&0&13&22&51\\
Split\_17&10&52&7&8&10&54&141&Split\_17&5&28&3&5&5&30&76\\
Split\_18&11&71&10&9&18&77&196&Split\_18&6&38&6&5&10&41&106\\
Split\_19&20&57&68&23&21&57&246&Split\_19&11&30&36&13&11&31&132\\
Split\_20&18&41&6&12&21&54&152&Split\_20&9&22&4&7&11&29&82\\
Split\_21&16&27&0&0&23&30&96&Split\_21&8&15&0&0&13&16&52\\
Split\_22&8&53&0&0&21&29&111&Split\_22&5&29&0&0&11&15&60\\
Split\_23&12&53&3&2&13&26&109&Split\_23&7&29&1&1&7&15&60\\
Split\_24&0&28&0&0&0&29&57&Split\_24&0&15&0&0&0&17&32\\
Split\_25&16&52&0&0&16&57&141&Split\_25&9&28&0&0&8&31&76\\
Split\_26&15&53&9&6&14&54&151&Split\_26&8&28&5&4&8&29&82\\
Split\_27&13&0&0&5&12&62&92&Split\_27&7&0&0&3&7&33&50\\
Split\_28&18&55&10&7&21&54&165&Split\_28&10&30&5&4&11&29&89\\
Split\_29&17&0&6&14&18&54&109&Split\_29&9&0&4&7&10&30&60\\
Split\_30&17&61&9&11&20&73&191&Split\_30&9&33&5&6&11&39&103\\
Split\_31&10&53&13&7&16&53&152&Split\_31&5&29&7&4&9&29&83\\
Split\_32&19&13&14&6&10&65&127&Split\_32&11&7&7&4&5&35&69\\
Split\_33&12&56&6&6&17&58&155&Split\_33&7&30&4&3&9&31&84\\
Split\_34&0&55&6&5&0&30&96&Split\_34&0&29&4&2&0&17&52\\
Split\_35&18&53&10&9&23&27&140&Split\_35&10&28&6&4&13&14&75\\

\bottomrule
\end{tabular}
}
\label{table:wisdm-data}
\end{table*}  
\begin{table*}[h]
\Huge
\centering
\caption{Class distribution of different training and test sets in the HHAR dataset. The correspondence between class indices and class names is as follows: 0-'bike', 1-'sit', 2-'stand', 3-'walk', 4-'stairs up', 5-'stairs down'.}
\renewcommand\arraystretch{1} 
\resizebox{0.7\linewidth}{!}{
\begin{tabular}{c|cccccc|c|c|cccccc|c}
\toprule
\multicolumn{16}{c}{\textbf{HHAR}} \\
\hline
\multirow{2}{*}{\textbf{Train}}&\multicolumn{6}{c|}{\textbf{Class Indices}}&\multirow{2}{*}{\textbf{Sum}}&\multirow{2}{*}{\textbf{Test}}&\multicolumn{6}{c|}{\textbf{Class Indices}}&\multirow{2}{*}{\textbf{Sum}}\\
\cline{2-7}\cline{10-15}
&0&1&2&3&4&5&&&0&1&2&3&4&5&\\
\hline
Split\_0&209&166&185&158&156&192&1066&Split\_0&89&71&79&68&67&83&457\\
Split\_1&269&176&187&208&177&232&1249&Split\_1&116&75&80&90&76&99&536\\
Split\_2&192&160&185&185&174&210&1106&Split\_2&83&69&79&80&74&90&475\\
Split\_3&179&160&164&189&152&220&1064&Split\_3&77&69&70&81&65&95&457\\
Split\_4&239&159&187&189&168&227&1169&Split\_4&102&68&81&81&72&97&501\\
Split\_5&257&164&193&206&168&216&1204&Split\_5&110&71&83&88&72&93&517\\
Split\_6&214&167&191&173&183&239&1167&Split\_6&91&72&82&74&79&103&501\\
Split\_7&192&165&209&190&172&188&1116&Split\_7&83&70&90&82&73&81&479\\
Split\_8&239&151&208&216&162&219&1195&Split\_8&103&65&89&92&70&94&513\\

\bottomrule
\end{tabular}
}
\label{table:hhar-data}
\end{table*}  
\begin{table*}[h]
\Huge
\centering
\caption{Class distribution of different training and test sets in the SSC dataset. The correspondence between class indices and class names is as follows: 0-`W', 1-`N1', 2-`N2', 3-`N3', 4-`REM'.}
\renewcommand\arraystretch{1} 
\resizebox{0.7\linewidth}{!}{
\begin{tabular}{c|ccccc|c|c|ccccc|c}
\toprule
\multicolumn{14}{c}{\textbf{SSC}} \\
\hline
\multirow{2}{*}{\textbf{Train}}&\multicolumn{5}{c|}{\textbf{Class Indices}}&\multirow{2}{*}{\textbf{Sum}}&\multirow{2}{*}{\textbf{Test}}&\multicolumn{5}{c|}{\textbf{Class Indices}}&\multirow{2}{*}{\textbf{Sum}}\\
\cline{2-6}\cline{9-13}
&0&1&2&3&4&&&0&1&2&3&4&\\
\hline
Split\_0&259&82&436&362&238&1377&Split\_0&112&35&187&155&102&591\\
Split\_1&223&141&855&141&242&1602&Split\_1&96&60&367&60&104&687\\
Split\_2&177&194&663&150&239&1423&Split\_2&76&84&284&64&103&611\\
Split\_3&193&74&619&132&286&1304&Split\_3&83&32&266&56&122&559\\
Split\_4&269&212&794&103&326&1704&Split\_4&116&91&340&44&140&731\\
Split\_5&301&111&583&174&173&1342&Split\_5&129&47&250&75&75&576\\
Split\_6&234&102&577&186&202&1301&Split\_6&101&44&247&79&87&558\\
Split\_7&372&121&556&269&256&1574&Split\_7&159&52&239&115&110&675\\
Split\_8&327&75&413&442&274&1531&Split\_8&140&32&178&190&117&657\\
Split\_9&203&70&750&194&348&1565&Split\_9&87&30&323&83&149&672\\
Split\_10&209&127&895&22&284&1537&Split\_10&90&55&383&9&122&659\\
Split\_11&176&22&629&168&216&1211&Split\_11&76&9&269&72&93&519\\
Split\_12&326&118&525&131&320&1420&Split\_12&140&51&225&56&137&609\\
Split\_13&108&40&348&103&120&719&Split\_13&47&17&149&44&52&309\\
Split\_14&258&39&553&207&312&1369&Split\_14&110&17&237&89&134&587\\
Split\_15&685&62&554&248&350&1899&Split\_15&294&26&238&107&150&815\\
Split\_16&276&68&635&205&318&1502&Split\_16&119&29&272&88&137&645\\
Split\_17&605&46&710&282&299&1942&Split\_17&260&19&305&121&128&833\\
Split\_18&199&126&474&355&164&1318&Split\_18&86&54&204&152&70&566\\
Split\_19&395&133&887&119&432&1966&Split\_19&169&57&380&51&186&843\\

\bottomrule
\end{tabular}
}
\label{table:ssc-data}
\end{table*}  
\begin{table*}[h]
\centering
\caption{Class distribution of different training and test sets in the MFD dataset. The correspondence between class indices and class names is as follows: 0-`Healthy', 1-`D1', 2-`D2'.}
\renewcommand\arraystretch{1} 
\resizebox{0.6\linewidth}{!}{
\begin{tabular}{c|ccc|c|c|ccc|c}
\toprule
\multicolumn{10}{c}{\textbf{MFD}} \\
\hline
\multirow{2}{*}{\textbf{Train}} & \multicolumn{3}{c|}{\textbf{Class Indices}} & \multirow{2}{*}{\textbf{Sum}} & \multirow{2}{*}{\textbf{Test}} & \multicolumn{3}{c|}{\textbf{Class Indices}} & \multirow{2}{*}{\textbf{Sum}} \\
\cline{2-4}\cline{7-9}
& 0 & 1 & 2 &&& 0 & 1 & 2 & \\
\hline
Split\_0 & 209 & 166 & 185 & 560 & Split\_0 & 89 & 71 & 79 & 239 \\
Split\_1 & 269 & 176 & 187 & 632 & Split\_1 & 116 & 75 & 80 & 271 \\
Split\_2 & 192 & 160 & 185 & 537 & Split\_2 & 83 & 69 & 79 & 231 \\
Split\_3 & 179 & 160 & 164 & 503 & Split\_3 & 77 & 69 & 70 & 216 \\
\bottomrule
\end{tabular}
}
\label{table:mfd-data}
\end{table*}

\subsection{The Statistical Details of the Dataset}
The statistical details of the dataset, such as the classes and domains are summarized in the following tables (\ref{table:ucihar-data},\ref{table:wisdm-data},\ref{table:hhar-data},\ref{table:ssc-data},\ref{table:mfd-data}) for ease of reference.

\begin{table*}[ht]
\caption{Down-Sampling Methods and Operations.}

\renewcommand\arraystretch{1.2} 
\begin{adjustbox}{width=1.0\textwidth,center}
\begin{tabular}{l|l|l}
\toprule
Method & Description & Operations \\ \hline
$+M_L$ & Dual convolutional down-sampling. & `Conv1d(C, 3, 2, 1, False)' \\
$+M_L+M_M$ & Max pooling followed by convolutional down-sampling. & `MaxPool1d(2)', `Conv1d(C, 3, 2, 1, False)` \\
$+M_M$ & Dual max pooling down-sampling. & `MaxPool1d(2)' twice \\
$+M_A$ & Dual average pooling down-sampling. & `AvgPool1d(2)' twice \\
$+M_R$ & Dual random-value down-sampling. & Random select, `MaxPool1d(2)' twice \\
\hline
\bottomrule
\end{tabular}
\end{adjustbox}
\label{table:down_sample}
\end{table*}

\subsection{The Details of the down-sampling Methods}
Parameters of the different down-sampling methods are summarized in Table~\ref{table:down_sample}.

\subsection{Domain Adaptation Loss}
A table summarizing the expressions of different \( L_d \) for various domain adaptation methods is provided in Table~\ref{table:formula}. 

\begin{table*}[ht]
\caption{For Deep CORAL, $\|\cdot\|_F^2$ denotes the squared Frobenius norm, $C_S$ and $C_T$ are covariance matrices of the source and target data. For HoMM, $n_s = n_t = \text{batch size}$, $\phi_\theta(\cdot)$ denotes the activation function, $\otimes$ denotes the outer product, and $p$ denotes the $p$-level tensor. For DSAN, $\mathcal{H}$ is the Reproducing Kernel Hilbert Space (RKHS) endowed with a characteristic kernel $k$. For DANN, $CE(\cdot)$ is the cross-entropy loss. For CDAN, $\tilde{Y}_{ST}$ is the output predicted with the fusion of $X_S$ and $X_T$, and $Y_{ST}$ is the concatenated label. For DIRT, $f_\theta(\cdot)$ is the embedding function. For CoDATS, $\mathcal{D}_{\text{KL}}$ is the Kullback-Leibler divergence. For AdvSKM and SASA, $\beta$ is the associative strength distribution.}

\renewcommand\arraystretch{1.2} 
\begin{adjustbox}{width=1.0\textwidth,center}
\begin{tabular}{l|l|c}
\toprule
Method & Loss Name & Domain Adaptation Loss \\ \hline
DDC & MMD loss & $\left\| \frac{1}{n_S}\sum_{x_s \in X_S}\phi(x_s) - \frac{1}{n_T}\sum_{x_t \in X_T}\phi(x_t) \right\|$ \\
\hline
Deep CORAL & Coral loss & $\frac{1}{4d^2} \| C_S - C_T \|_F^2$ \\
\hline
HoMM & High-order MMD loss & $\frac{1}{L^p} \left\| \frac{1}{n_s}\sum_{i=1}^{n_s}\phi_\theta(x_s^i)^{\otimes p} - \frac{1}{n_t}\sum_{i=1}^{n_t}\phi_\theta(x_t^i)^{\otimes p} \right\|_F^2$ \\
\hline
MMDA & MMD loss & $\left\| \frac{1}{n_S}\sum_{x_s \in X_S}\phi(x_s) - \frac{1}{n_T}\sum_{x_t \in X_T}\phi(x_t) \right\|$ \\
& Coral loss & $\frac{1}{4d^2} \| C_S - C_T \|_F^2$ \\
\hline
DSAN & Local MMD loss & $\mathbb{E}_c \left\| \mathbb{E}_{p(c)}[\phi(X_S)] - \mathbb{E}_{q(c)}[\phi(X_T)] \right\|_{\mathcal{H}}^2$ \\
\hline
DANN & Domain loss & $CE(\hat{Y}_S, Y_S) + CE(\hat{Y}_T, Y_T)$ \\
\hline
CDAN & Domain loss & $CE(\tilde{Y}_{ST}, Y_{ST})$ \\
\hline
DIRT & Domain loss & $\sup_D \left[ \mathbb{E}_{x \sim D_S} \ln D(f_\theta(x)) + \mathbb{E}_{x \sim D_T} \ln (1 - D(f_\theta(x))) \right]$ \\
\hline
CoDATS & Adversarial loss & $CE(F(X_S), Y_S) + \mathbb{E}_{X \sim D_T^X} \left[ CE(F(X_T), Y_T) + \mathcal{D}_{\text{KL}}(Y_{\text{true}} \| \mathbb{E}_{X \sim D_T^X}[C(F(X))]) \right]$ \\
\hline
AdvSKM & Adversarial MMD loss & $\sum_{i,j=1}^{N_S} \frac{K(\phi(x^s_i), \phi(x^s_j))}{N_S^2} + \sum_{i,j=1}^{N_T} \frac{K(\phi(x^t_i), \phi(x^t_j))}{N_T^2} - 2\sum_{i=1}^{N_S}\sum_{j=1}^{N_T} \frac{K(\phi(x^s_i), \phi(x^t_j))}{N_SN_T}$ \\
\hline
SASA & Domain loss & $\sum_{m=1}^{M} \left\| \frac{1}{n_S}\sum_{x_S \in X_S} \beta_T^m - \frac{1}{n_T}\sum_{x_T \in X_T} \beta_T^m \right\|$ \\
\hline
\bottomrule
\end{tabular}
\end{adjustbox}
\label{table:formula}
\end{table*}

\subsection{Limitations}
While the proposed method demonstrates good performance and interpretability, several limitations remain. First, as shown in Table~\ref{table:FLOPs}, the computational cost of the model—particularly during training—may be significant due to the complexity of the architecture and the need for uncertainty estimation. This may limit its applicability in resource-constrained environments or real-time settings. Exploring more efficient approximations or lightweight model variants could help mitigate this overhead.
Secondly, the methods can be further evaluated in more challenging real-world scenarios, such as variations across different scanners, institutions, and patient populations.

\begin{table*}[h]
\Huge
\centering
\caption{The Floating Point Operations (FLOPs) of different methods measured at a batch size of 32.}
\renewcommand\arraystretch{1} 
\resizebox{0.8\linewidth}{!}{
\begin{tabular}{l|cccccccccccc}
\toprule
\multicolumn{13}{c}{\textbf{FLOPs (M)}} \\
\hline
&noAdapt&DDC&Deep&HoMM&DANN&MMDA&DSAN&CDAN&DIRT&CoDATS&AdvSKM&SASA\\
\hline
baseline&29.20&58.39&58.39&58.39&58.47&58.39&58.39&59.07&58.55&59.48&59.18&58.70 \\
$+UN$&29.20&58.39&58.39&58.39&58.47&58.39&58.39&59.07&58.55&59.48&59.18&58.70 \\
$+UN+M_L$&53.39&106.78&106.78&106.79&107.09&106.78&106.79&108.49&107.42&115.20&108.35&108.04 \\
$+UN+M_L+M_M$&53.37&106.74&106.74&106.75&107.06&106.74&106.75&108.46&107.38&115.16&108.32&108.00 \\
$+UN+M_R$&53.36&106.71&106.71&106.72&107.03&106.71&106.72&108.43&107.35&115.13&108.29&107.97 \\
$+UN+M_M$&53.36&106.71&106.71&106.72&107.03&106.71&106.72&108.43&107.35&115.13&108.29&107.97 \\
$+UN+M_A$&53.36&106.71&106.71&106.72&107.03&106.71&106.72&108.43&107.35&115.13&108.29&107.97 \\
\bottomrule
\end{tabular}
}
\label{table:FLOPs}
\end{table*}

\ifCLASSOPTIONcaptionsoff
  \newpage
\fi

\bibliographystyle{IEEEtran}
\bibliography{egbib}

\end{document}